\documentclass[10pt,twocolumn,letterpaper]{article}

\usepackage[pagenumbers]{cvpr} 

\pdfoutput=1
\usepackage{graphicx}
\usepackage{amsmath}
\usepackage{amssymb}
\usepackage{booktabs}
\usepackage{subcaption}

\usepackage{url}
\usepackage{graphicx}
\usepackage{booktabs}
\usepackage{adjustbox}
\usepackage{xcolor}
\usepackage{algorithm}
\usepackage{algpseudocode}
\usepackage{amsmath}
\usepackage{amssymb}
\usepackage{bm}
\usepackage{caption}
\usepackage{wrapfig}
\usepackage{multirow}
\usepackage{threeparttable}
\usepackage{bbding}
\newcommand{\indep}{\perp \!\!\! \perp}

\newtheorem{definition}{Definition}

\DeclareMathOperator*{\argmin}{argmin}
\newcommand{\ood}{{o.o.d}\onedot}
\newcommand{\Ood}{{O.o.d}\onedot}

\usepackage[capitalize]{cleveref}
\crefname{section}{Sec.}{Secs.}
\Crefname{section}{Section}{Sections}
\Crefname{table}{Table}{Tables}
\crefname{table}{Tab.}{Tabs.}

\begin{document}

\title{Direct-Effect Risk Minimization for Domain Generalization}

\author{Yuhui Li\thanks{Equal contribution.} \\
Peking University\\
{\tt\small yuhui.li@stu.pku.edu.cn}
\and
Zejia Wu\footnotemark[1] \thanks{Work done during an internship at University of Waterloo.}\\
University of Waterloo\\
{\tt\small zejia.woo@gmail.com}
\and
Chao Zhang \\
Peking University\\
{\tt\small c.zhang@pku.edu.cn}
\and
Hongyang Zhang \\
University of Waterloo \\
{\tt\small hongyang.zhang@uwaterloo.ca}
}

\maketitle

\begin{abstract}
   We study the problem of out-of-distribution (\ood) generalization where spurious correlations of attributes vary across training and test domains. This is known as the problem of correlation shift and has posed concerns on the reliability of machine learning. In this work, we introduce the concepts of direct and indirect effects from causal inference to the domain generalization problem. {We argue that models that learn direct effects minimize the worst-case risk across correlation-shifted domains.} To eliminate the indirect effects, our algorithm consists of two stages: in the first stage, we learn an indirect-effect representation by minimizing the prediction error of domain labels using the representation and the class labels; in the second stage, we remove the indirect effects learned in the first stage by matching each data with another data of similar indirect-effect representation but of different class labels in the training and validation phase. Our approach is shown to be compatible with existing methods and improve the generalization performance of them on correlation-shifted datasets. Experiments on 5 correlation-shifted datasets and the DomainBed benchmark verify the effectiveness of our approach. Our code is available at 
   {\url{https://github.com/Liyuhui-12/DRMforDG}}.
\end{abstract}

\section{Introduction}

\begin{figure}
    \centering
    \includegraphics[width=1\linewidth]{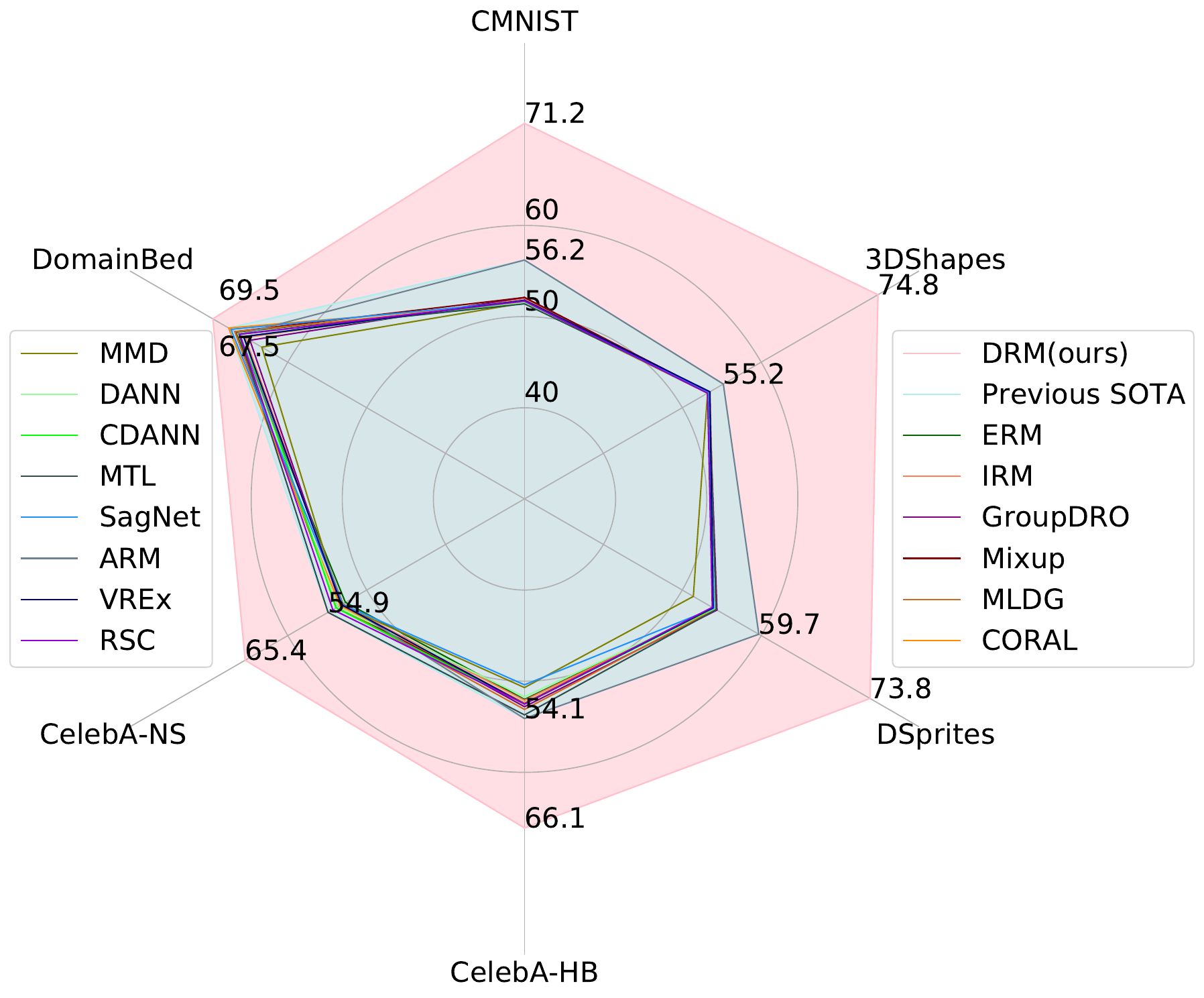}
    \vspace{-0.3cm}
    \caption{Test accuracy of \ood algorithms on 5 correlation-shifted datasets and the DomainBed benchmark (avg). The pink region represents the performance of our method, while the light blue region represents the previously best-known results (implemented by DomainBed using \emph{training-domain validation}) on each dataset.}
    \label{figure: performance}
    \vspace{-0.6cm}
\end{figure}

Machine learning has achieved huge success in many fields, yet they mostly rely on the independent and identically distributed (\iid) assumption. When it comes to an out-of-distribution (\ood) test domain, machine learning models usually suffer from a sharp performance drop \cite{beery2018recognition,arjovsky2019invariant,nagarajan2020understanding}. The \ood data typically come in the form of \emph{correlation shift}, where spurious correlations of attributes vary between training and test domains, or \emph{diversity shift}, where the shifted test distribution keeps the semantic content of the data unchanged while altering the data style \cite{ye2022ood}. The focus of this work is on the former setting known as correlation shift. That is, given stable causality and spurious correlations between attributes, how to disentangle the stable causality and the spurious correlations from the training data. Figure \ref{figure: performance} shows the performance gain of our method on the correlation shift datasets.

\begin{figure*}[t]
    \centering
    \includegraphics[width=1.0\linewidth]{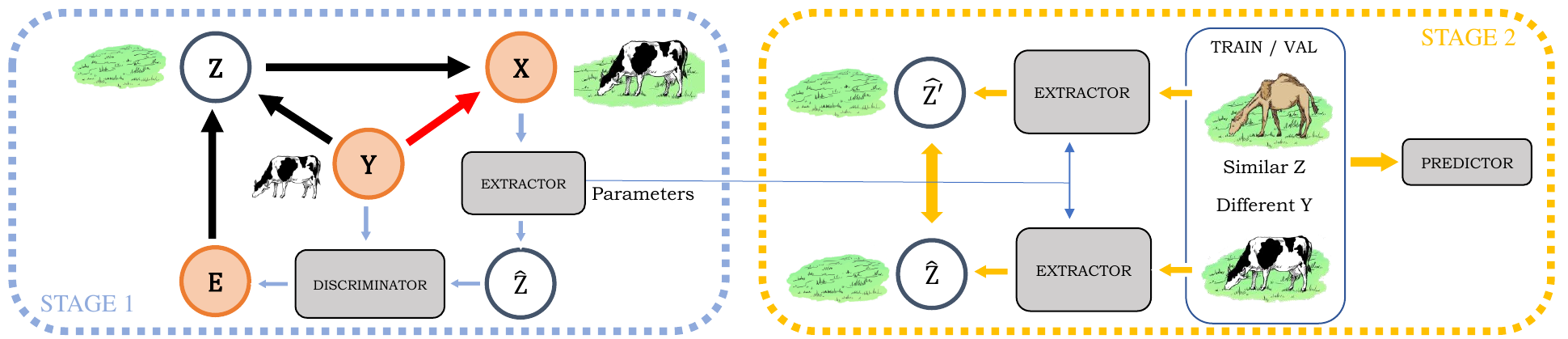}
    \vspace{-0.5cm}
    \caption{Description of our two-stage approach. In \textcolor[RGB]{143,170,220}{Stage 1}, we jointly learn a discriminator and an (indirect-effect) extractor by predicting the domain labels. In \textcolor[RGB]{255,192,0}{Stage 2}, the extractor in Stage 1 is used to construct a balanced batch of samples with a similar indirect-effect representation but different class labels, as well as a balanced validation set which have no access to the test domain. A predictor is trained on the balanced batch and validated on the balanced validation set to predict the class labels. The flow of our method is represented by the \textcolor[RGB]{143,170,220}{blue} and the \textcolor[RGB]{255,192,0}{yellow} arrows. The red and black arrows form the graphical model of the correlation shift problem (data generation process). The red arrows represent the direct effect and the black arrows represent the indirect effect.}
    \label{causal_graph}
\vspace{-0.5cm}
\end{figure*}

Much effort has been devoted to learning representations that are invariant across training environments, where many works have introduced the tools from causality to address the \ood generalization problems. When the data are of high dimension and multiple attributes are entangled, it is challenging to identify invariant causality  across domains. Many methods have been designed to resolve the issue. Representative methods include incorporating invariance constraints by designing new loss functions \cite{arjovsky2019invariant,krueger2021out,bellot2020accounting,lin2022bayesian}, learning latent semantic features in causal graphs by VAE \cite{liu2021learning,lu2021invariant}, and eliminating selection bias by matching \cite{mahajan2021domain,wang2022causal}. However, these methods, despite their theoretical guarantees, fail to show empirical improvement over Empirical Risk Minimization (ERM) as verified by the DomainBed benchmark \cite{gulrajani2020search,vedantam2021empirical}.

This paper is the first attempt to use the tool of \emph{direct} and \emph{indirect effects} from causal inference to analyze the correlation shift problem. We argue that under certain conditions, models that learn direct effects minimize the worst-case risk across domain-shifted domains. To learn the direct effects, we propose a two-stage approach: in the first stage, we use an extractor to infer the indirect-effect representation $Z$ from the data $X$ such that $Z$ can predict the domain label $E$ through a discriminator head (see the blue box in Figure \ref{causal_graph}). In the second stage, we construct a \emph{balanced batch} by augmenting the original training batch and validation set with data of the same indirect-effect representation $Z$ but of a different class label $Y$. We demonstrate that our validation balancing approach can overcome the inconsistency of the validation distribution with the test distribution, which was shown to be an important reason for the performance degradation of many existing approaches under the DomainBed protocol. We test our approach on the DomainBed benchmark. On the correlation shift dataset \emph{Colored MNIST}, our model obtains an average accuracy of $71.2\%$ over three domain generalization problems. While the information-theoretic best accuracy on the \emph{Colored MNIST} dataset is $75\%$, our method achieves an accuracy as high as $69.7\%$ in the most difficult ``$-90\%$'' environment. Moreover, we provide evidence that the foundation models can alleviate the diversity shift problem but cannot solve the correlation shift problem well, demonstrating that our approach closes the gap between foundation models and domain generalization to some extent. Our method can be combined with existing domain generalization methods to significantly improve their performance on correlation shift datasets.
Our main contributions are as follows:
\begin{itemize}
\item
We present a framework to analyze the correlation shift problem based on direct/indirect causal effects.
\item
We propose a new approach to improve \ood generalization. We recover the indirect-effect representation and eliminate the indirect effect during training and validation. We also show that our model selection method can largely overcome the model selection problem caused by the inconsistency between the validation and the test distribution. Our approach can be easily compatible with other algorithms and substantially improve their performances.
\item
Our method outperforms baselines by a large margin on the correlation-shifted datasets. For example, our approach achieves up to 15\% absolute improvement on the \emph{Colored MNIST} dataset and up to 11\% absolute improvement on the \emph{CelebA} datasets over the state-of-the-art in terms of average accuracy over three domains.
\end{itemize}

\section{Preliminaries}

\textbf{Notations.} In this paper, we will use \emph{capital} letters such as $X$, $Y$, and $Z$ to represent random variables, \emph{lower-case} letters such as $x$ and $z$ to represent realization of random variables, and letters with \emph{hat} such as $\hat{Z}$ to represent inferred variables by the model. We use the \emph{calligraphic capital} letter $\mathcal{E}$ to represent the set of environments and by \emph{lower-case} letter $e$ the domain label. $X \indep Y$ means that random variables $X$ and $Y$ are independent. We use $\mathcal{D}^{e}$ and $\mathbb{P}^{e}$ to denote the distribution and the corresponding probability density function (PDF) in environment $e$, respectively. We add the superscript $e$ to a variable such as $x^{e}$ to indicate that the variable is sampled from the distribution of the environment $e$, and $(x^{e}_{i},y^{e}_{i},e)$ refers to an instance sampled from $\mathcal{D}^{e}$. We denote by $\mathcal{H}$ the hypothesis class of models and by $h:\mathcal{X}\rightarrow\mathcal{Y}$ the predictor. $R^{e}(h)$ refers to the risk of predictor $h$ in the environment $e$. \emph{Environment} and \emph{domain} are of the same concept, and we use them interchangeably throughout the paper.

\subsection{Correlation shift}

We consider the correlation shift problem in this paper. In a supervised learning setting, the goal is to learn the labeling function $f: \mathcal{X}\rightarrow\mathcal{Y}$, which is consistent in all environments. However, there often exists a variable set $\mathcal{Z}$ such that there are spurious correlations between $Z \in \mathcal{Z}$ and the label $Y$. When the spurious correlation changes with the environment, the model that utilizes the spurious correlation may face a performance breakdown in the new test environment. Spurious correlations may originate from the data generation process or selection bias, which is very common in reality. 

\begin{figure}[t]
    \centering
    \includegraphics[width=0.7\linewidth]{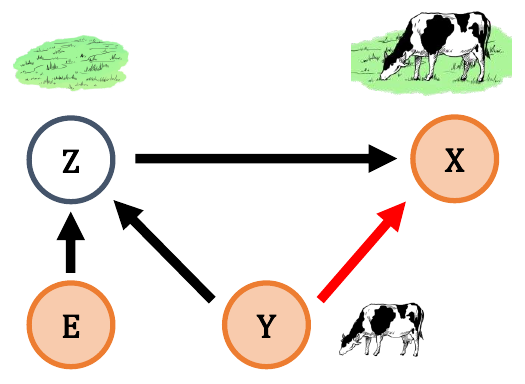}
    \caption{The causal graph of correlation shift problems. {The red arrow represents the stable direct effect, while the black pathways represent the indirect effects that changes as the environment changes, which is what we want to remove.}}
    \label{cg}
\vspace{-0.5cm}
\end{figure}

Consider a binary classification problem of cows and camels (see Figure \ref{cg}). We assume that the animal category and background are the two attributes that contribute to the generation of an image. Our goal is to predict the animal category $Y$ from image $X$, and the background is denoted by $Z$. The image $X$ is the result of the total effect of the two attributes. We assume that the value of $Y$ is changed from ``cow'' to ``camel'' during the data generation process. So the animal in the image $X$ is changed. Meanwhile, the cow is more likely to be on the grass, while the camel is more likely to be in the desert. Therefore, $Z$ may change from ``grass'' to ``desert'' as $Y$ changes, changing the background in the image $X$. We define the correlation shift as follow, which is consistent with that in \cite{ye2022ood}.
\begin{definition}[Correlation shift]
Assume that we have a training environment $e_{S} \in \mathcal{E}_{train}$ and a test environment $e_{T} \in \mathcal{E}_{test}$, whose probability density functions are $\mathbb{P}^{e_{S}}$ and $\mathbb{P}^{e_{T}}$, respectively. Assume that $\mathbb{P}^{e_{S}}(y)=\mathbb{P}^{e_{T}}(y)$ for every $y \in \mathcal{Y}$, and that $e_{S}$ and $e_{T}$ share the same labeling function. Then there exists correlation shift between $e_{S}$ and $e_{T}$ if there exists a set $\mathcal{Z}$ such that
\begin{equation*}
    \sum_{y\in\mathcal{Y}}\int_{\mathcal{Z}} \lvert \mathbb{P}^{e_{S}}(z|y)-\mathbb{P}^{e_{T}}(z|y) \rvert dz \neq 0,
\end{equation*}
where $\mathbb{P}^{e_{S}}(z) \times \mathbb{P}^{e_{T}}(z) \neq 0$.
\end{definition}

By definition, we consider a direct acyclic graph (DAG) describing the data generation process (see Figure \ref{cg}), where the pathways from $Y$ to $X$ are composed of two parts: $Y \rightarrow X$ and $Y \rightarrow Z \rightarrow X$. In causal inference, the former is referred to as the \emph{direct effect} of $Y$ on $X$, while the latter is referred to as the \emph{indirect effect}. $Z$ is a child of the environment $E$, which leads to the indirect effects varying with the environment. In a supervised learning setting, models are trained to learn the reversed process of the data generation process, \ie $X \rightarrow Y$ or $X \rightarrow Z \rightarrow Y$. To obtain models that can generalize across different domains under correlation shift, we need to cut off the indirect effect pathway ($Y \rightarrow Z \rightarrow X$) and force the model to learn the reversed mapping of the robust direct effects ($Y \rightarrow X$).

{
It is worth noting that in fact there is often only a correlation between the label $Y$ and the mediator $Z$. For example, the presence of a cow does not lead to the presence of grass. For the model, however, it is reasonable to interpret this correlation as an indirect causal effect in the absence of human knowledge. Under this interpretation, all confounders between $Y$ and $X$ can be included by $Z$, thus there are no unblocked backdoor pathways \cite{pearl2009causality} between $Y$ and $X$. Our description of the data generation process under correlation shifts and the DAG are general in nature and consistent with other recent works like \cite{wang2022causal}. 
}

\subsection{Problem Setting}
We consider a standard domain generalization setting, where the data come from different environments $e \in \mathcal{E}_{all}$. Assume that we have the training data collected from a finite subset of training environments $\mathcal{E}_{train}$, where $\mathcal{E}_{train} \subset \mathcal{E}_{all}$. For every environment $e \in \mathcal{E}_{train}$, the training dataset $\left\{\left(x_{i}^{e}, y_{i}^{e}, e\right)\right\}_{i=1}^{N_{e}}$ is sampled from the distribution $\mathcal{D}^{e}$. The PDF of the distribution is $\mathbb{P}^e(X^e,Y^e)=\mathbb{P}(X,Y\mid E=e)$, where $X$ is the instance (\eg, an image), $Y$ is the class label, $E$ is the domain label, and $N_e$ is the number of training data in environment $e$. The goal of domain generalization is to train a model with data from training environments $\mathcal{E}_{train}$ that generalizes well to all environments $e \in \mathcal{E}_{all}$. Our goal is to find a predictor $h^*:\mathcal{X} \rightarrow \mathcal{Y}$ in the hypothesis class $\mathcal{H}$ such that the worst-case risk is minimized:
\begin{equation}
\label{def_of_ood}
h^*=\argmin _{h \in \mathcal{H}} \max _{e \in \mathcal{E}_{\text {all }}} R^{e}(h),
\end{equation}
where $R^{e}(h)$ is the risk of predictor $h$ in environment $e$. {We argue that the model learning the stable direct effects is robust when the environment changes, which satisfies equation \ref{def_of_ood}.} To enable the model to learn the direct effects in the data, it is desirable to cut off the pathway between $Z$ and $Y$ so that they are independent. To this end, we designed a novel framework with improved training process and model selection.

\section{Method}
\subsection{Recovering Indirect Effects}
\label{Recover IE}
Since the variable $Z$ on the indirect-effect pathway is often not observable, we design an extractor to recover the representation $\hat{Z}$ of the indirect effect from $X$ by learning a discriminator head in the first stage. {From Figure \ref{cg}, we observe that the indirect-effect representation $Z$ and the class label $Y$ form a Markov blanket for the domain label $E$, which means that $E$ is independent of other variables given $Y$ and $Z$.} Hence the discriminator head needs $Z$ and $Y$ to predict $E$. If we take the output of the extractor and $Y$ as the input of the discriminator head, the discriminator head will force the extractor to recover $Z$ from $X$. Specifically, assume that the dataset is sampled from $N_{S}$ training domains. We set up an extractor $G(\cdot; \Theta_{G}):\mathcal{X} \rightarrow \mathcal{Z}$ and a discriminator head $D(\cdot, \cdot; \Theta_{D}):\mathcal{Z} \times \mathcal{Y} \rightarrow [0,1]^{N_{S}}$ that outputs the probability that a sample belongs to each training domain, and update the parameters of both models by minimizing the prediction error of domain label $e$:

\begin{equation}
\label{1stage}
\begin{aligned}
    \Theta_{G}^{*},\Theta_{D}^{*} &:= \\
    &\argmin_{\Theta_{G},\Theta_{D}}\mathbb{E}_{x,y,e}\textrm{CE}(D(G(x;\Theta_{G}),y;\Theta_{D}), e),
\end{aligned}
\end{equation}
where $\mathrm{CE}$ is the Cross Entropy loss, $\Theta_{G}$ and $\Theta_{D}$ stand for the parameters of the extractor $G$ and the discriminator head $D$, respectively, and $(x,y,e)$ is a training sample. We use the learned extractor to obtain the representation $\hat{z}_{i}^{e}=G(x_{i}^{e};\Theta_{G}^{*})$ for every instance $(x_{i}^{e},y_{i}^{e},e)$. 

Many methods learned domain discriminators by a minimax problem \cite{ganin2016domain,li2018domain,albuquerque2019generalizing}. These methods extracted features that could maximize the domain discriminator error. In our approach, on the other hand, the representation vector $Z$ is obtained by minimizing the domain discrimination error. This makes our model easier to optimize and more stable than a minimax game.

\subsection{Eliminating Indirect Effects in Training (TB)}
\label{stage2}
In the model training stage, we remove the indirect effects from the data by creating balanced batches based on the representation $\hat{Z}$, which is referred to as \textbf{TB} in the following. We start by defining the balanced batch.

\begin{definition}[Balanced Batch]
\label{balanced_batch}
For any sample in a balanced batch, denoted by $(x_{i}^{e},y_{i}^{e},e,\hat{z}_{i}^{e})$, there exists a corresponding sample $(x_{j}^{e},y_{j}^{e},e,\hat{z}_{j}^{e})$ with probability $P$, such that $\hat{z}_{i}^{e}=\hat{z}_{j}^{e}$, $y_{i}^{e} \neq y_{j}^{e}$, and $\mathbb{P}_{Batch}(Y)=\mathbb{P}_{D}(Y)$, where $\mathbb{P}_{Batch}(Y)$ and $\mathbb{P}_{D}(Y)$ are marginal probability density functions of $Y$ in the batch and the training set, respectively. 
\vspace{-0.2cm}
\end{definition}
Ideally, for each sample $x_{i}$, we can find a corresponding sample $x_{j}$ with the same indirect-effect representation $\hat{z}_{i}^{e}=\hat{z}_{j}^{e}$. However, we cannot always find exactly equal $\hat{z}$ as in the ideal case. To resolve this problem, for each sample $(x_{i}^{e},y_{i}^{e},e,\hat{z}_{i}^{e})$, we search for another sample $(x_{j}^{e},y_{j}^{e},e,\hat{z}_{j}^{e})$ such that $\hat{z}_{j}^{e}$ is the nearest neighbor of $\hat{z}_{i}^{e}$. To ensure that the marginal distribution of label $Y$ does not change, we include the matched sample into the batch with a probability that depends on the proportion of each class of samples in the training set.

Taking the cow and camel classification problem in Figure \ref{cg} as an example, we search for a camel image with the same background for each cow image in the batch, \eg, a camel standing on grass for a cow standing on grass, as well as a cow image with the same background for each camel image. Thus our training batches consist of pairs of images. We train the model on balanced batches constructed as described above.

\begin{algorithm}[t]
  \caption{Direct-Effect Risk Minimization (DRM)}
  \label{alg1}
  \textbf{Input:} Training set $D$; validation set $V$; initial predictor $f_{\theta_0}$; training steps $T$; checkpoint frequency $C$; learning rate $\epsilon$;\\
  \textbf{Output:} 
      Predictor $f_{\theta_T}$;
  \begin{algorithmic}[1]
    \State  Update $\Theta_{G},\Theta_{D}$ by the equation \ref{1stage} and get $\Theta_{G}^{*},\Theta_{D}^{*}$;
        \State $V_b \gets \{\}$;
        \For{$(x^{e},y^{e})$ in $V$}
            \State $\hat{z}^{e} \gets G(x^{e};\Theta_{G}^{*})$;
            \State Search $V$ for $(x^{e}_b,y^{e}_b)$ with the closest $\hat{z}^{e}_{b}$ to $\hat{z}^{e}$, and $y^{e} \neq y^{e}_b$;
            \State Add $(x^{e},y^{e})$, $(x^{e}_b,y^{e}_b)$ to $V_b$;
        \EndFor
	\State $t \gets 0$;
    \While{$t\le T$}
            \State Sample a batch $B = \left\{\left(x_i^e,y_i^e\right)\right\}_{i=1}^{\text{batchsize}}$ from $D$; 
    		\For{$(x^{e},y^{e})$ in batch}
    			\State $\hat{z}^{e} \gets G(x^{e};\Theta_{G}^{*})$;
    			\State Search $D$ for $(x^{e}_b,y^{e}_b)$ with the closest $\hat{z}^{e}_{b}$ to $\hat{z}^{e}$,
                           and $y^{e} \neq y^{e}_{b}$;
    			\State Add $(x^{e}_b,y^{e}_b)$ to $B$;
    		\EndFor
    		\State Run ERM or other algorithms on $B$ and update $f_{\theta_t}$;
                \If{$C|t$}
                    \State Evaluate model on balanced validation set $V_{b}$;
                \EndIf
    		\State $t\leftarrow t+1$;
    \EndWhile
\end{algorithmic}
\end{algorithm}

\subsection{Model Selection (VB)}
\begin{figure}[ht]
	\centering  
	\subcaptionbox{IRM}{
		\includegraphics[width=0.31\linewidth]{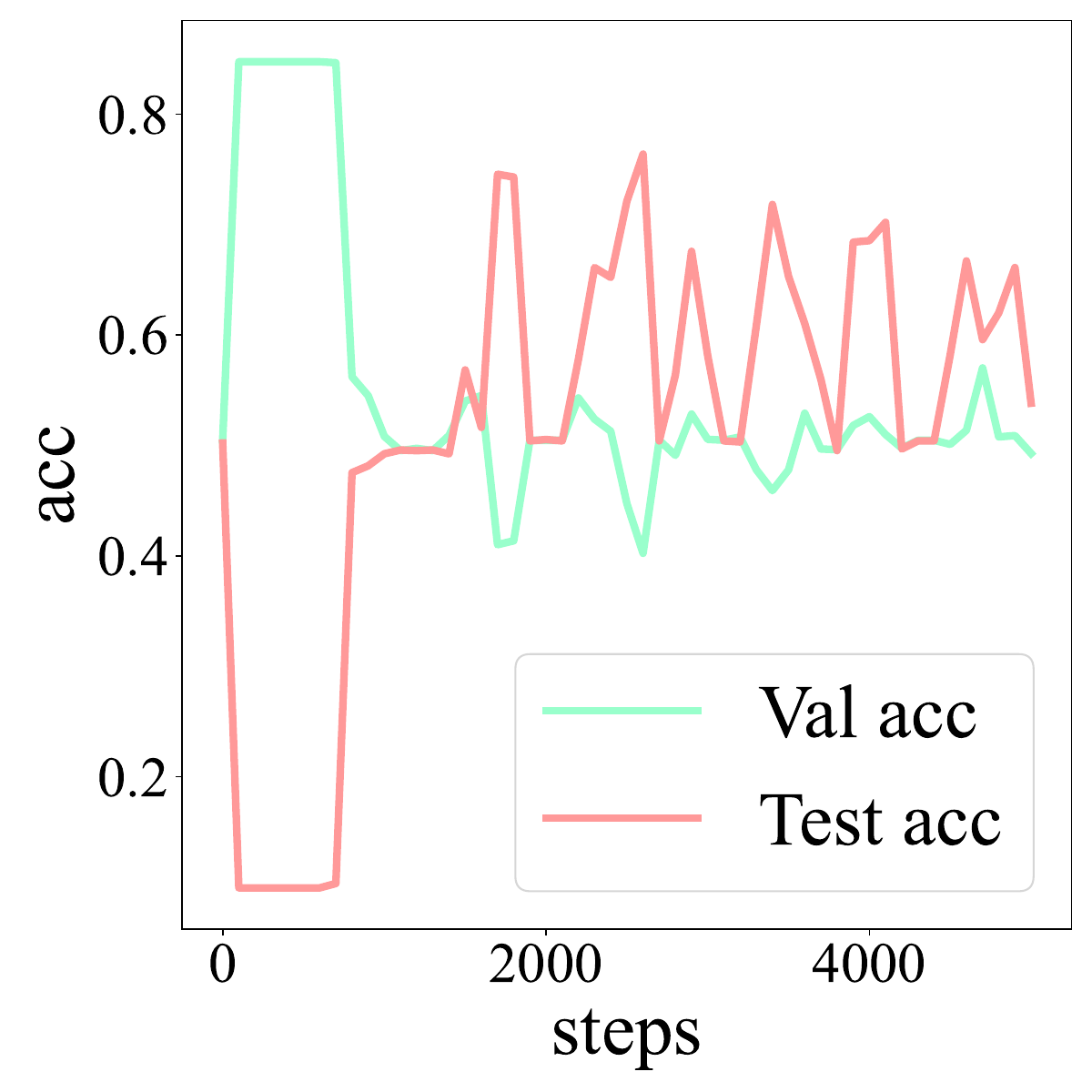}}
	\subcaptionbox{VREx}{
		\includegraphics[width=0.31\linewidth]{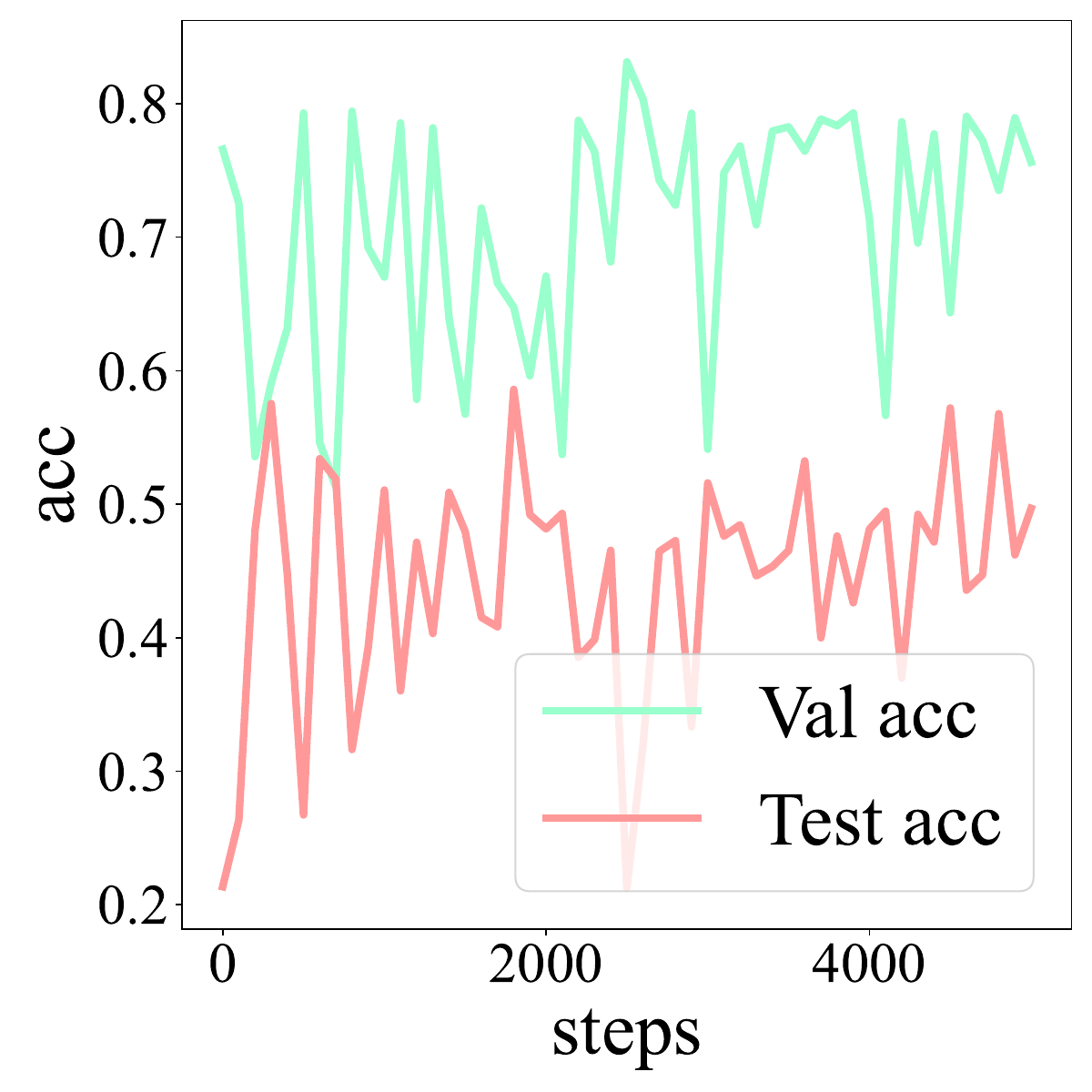}}
	\subcaptionbox{CORAL}{
		\includegraphics[width=0.31\linewidth]{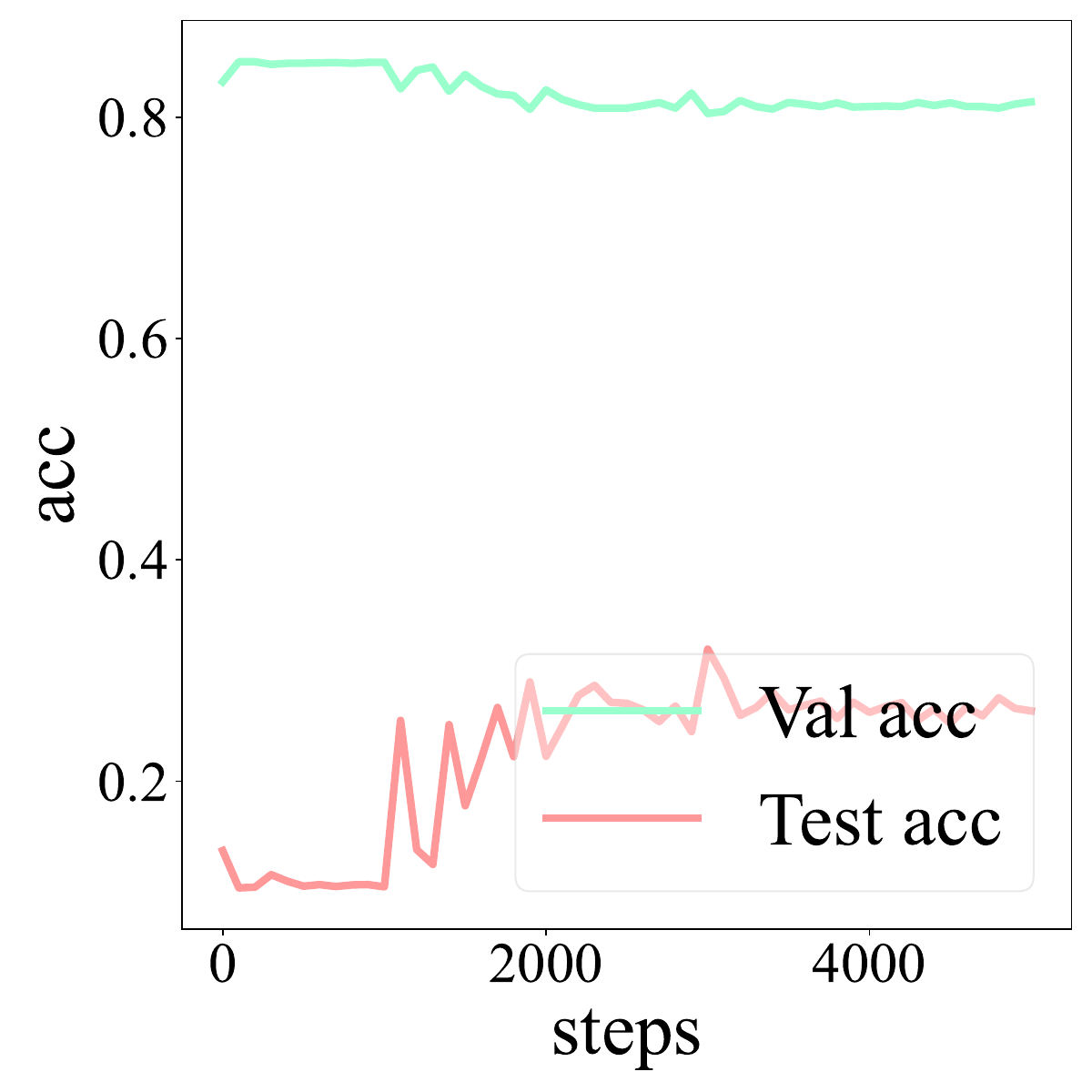}}
	\caption{The inconsistency of validation set accuracy and test set accuracy during the training process. (a) is for IRM, (b) is for VREx, and (c) is for CORAL.}
	\label{DomainBedfig0}
\end{figure}

According to DomainBed\cite{gulrajani2020search}, whether to follow the protocol of random hyperparameter search can drastically affect the performance of a method, especially for correlation-shifted datasets. DomainBed recommended that researchers should disclaim any oracle-selection results as such and specify policies to limit access to the test domain. Unlike the \iid task, the distributions of the training and testing domains are significantly different in a domain generalization problem, and there is a large performance gap between selecting a model on the test domain distribution and the training domain distribution. For example, as shown in Figure \ref{DomainBedfig0}, while using the method of training-domain validation, the validation and test accuracy rates are often inconsistent. Checkpoints with high accuracy on the validation set do not perform well on the test set. DomainBed searches for hyperparameters randomly to ensure no access to the test domain, thus the results of many existing methods reported by DomainBed may be lower. We believe that one of the very significant reasons for this decline in performance is that the spurious correlation in the validation set misleads the model selection. Models that utilize spurious correlation are able to perform well on validation sets where spurious correlation exists, while they cannot generalize to unseen test domains with reversed correlation.

Considering this inconsistency, our DRM framework also includes a novel approach to model selection based on the above-mentioned balancing approach, which is referred to as \textbf{VB} in the following sections. Specifically, we create balanced validation sets in the same way as in Section \ref{Recover IE} and Section \ref{stage2}, on which we evaluate the models. The validation data are all divided from the training domains, which is consistent with the training-domain validation protocol in DomainBed and ensures that the model has no access to the test set.

For the main results of our paper (Table \ref{table_corr}), we run ERM on balanced batches and evaluate the models on balanced validation sets (ERM+VB+TB). However, it is worth mentioning that our VB and TB methods can also be combined with many existing methods and substantially improve their performances. We show these results in Table \ref{ab_cmnist}. All of our experiments followed the DomainBed protocol, and to be fair, we only compared DRM with methods that follow the same protocol and have no access to test domains.

The pseudo-code description of the whole DRM framework is shown in Algorithm \ref{alg1}.

\section{Experiments}

We compare DRM with 14 baseline methods, including: ERM~\cite{vapnik1998statistical}, IRM~\cite{arjovsky2019invariant}, GroupDRO~\cite{sagawa2019distributionally}, Mixup~\cite{zhang2018mixup,xu2020adversarial,yan2020improve,wang2020heterogeneous}, MLDG~\cite{li2018learning}, CORAL~\cite{sun2016deep}, MMD~\cite{li2018mmd}, DANN~\cite{ganin2016domain}, CDANN~\cite{li2018deep}, MTL~\cite{blanchard2011generalizing}, SagNet~\cite{nam2019reducing}, ARM~\cite{zhang2020adaptive}, VREx~\cite{krueger2021out} and RSC~\cite{huang2020self}, which appeared in the DomainBed benchmark \cite{gulrajani2020search}. We evaluate the performance of our approach on datasets classified as correlation-shifted datasets by \cite{wiles2021fine, ye2022ood}, in which there are spurious correlations between the class label and the features such as the color or the background of the images. Learning the test environment is difficult since there might be a spurious-correlation flip between the training and test environments. We show that the performance of \iid algorithms such as ERM will significantly drop in this case, while our approach achieves improved performance in these difficult environments. We strictly follow the protocol of DomainBed by conducting random searches for all hyperparameters in all stages. For comparison, we use the codes provided by DomainBed to run the above-mentioned 14 methods. We defer the details of the experiment setting and results to Appendix \ref{app_exp}.

\subsection{Datasets}

We evaluate our approach on \emph{CMNIST} dataset, \emph{3DShapes} dataset, \emph{DSprites} dataset, and \emph{CelebA} dataset, which are common correlation shift datasets used to evaluate domain generalization methods \cite{wiles2021fine,ye2022ood,sagawa2019distributionally}.

\emph{Colored MNIST}\cite{arjovsky2019invariant, lecun1998mnist} creates a spurious correlation between colors and digits by artificially coloring the digits red or green. The correlations between colors and labels in three environments are ${+90\%}$, ${+80\%}$, and ${-90\%}$, respectively. For example, in the ``${+90\%}$" environment, ${90\%}$ images with label $1$ are dyed red, while ${90\%}$ images with label $0$ are dyed green. In addition, the dataset randomly flips ${25\%}$ of the class labels, which results in ${75\%}$ correlation between shapes and digital labels, lower than that between colors and labels. Thus, an \iid learning approach like ERM prefers to learn correlations between colors and labels. To test the validity of our method in a broader and more realistic context, we also introduced four correlation shift datasets, which are \emph{3DShapes}\cite{3dshapes18}, \emph{DSprites}\cite{dsprites17}, \emph{CelebA-HB}, and \emph{CelebA-NS}\cite{liu2015faceattributes}. The stable and spurious features for these datasets are ``floor hue'' and ``orientation'', ``Position X'' and ``Position Y'', ``No Beard'' and ``Wearing Hat'', ``Smiling'' and ``Wearing Necktie'', respectively. We defer the specific details about the dataset to Appendix \ref{ds}.

While focusing on the correlation shift problem, we also conduct experiments on the diversity-shifted dataset contained in DomainBed such as \emph{PACS} and \emph{VLCS} to show that DRM will not hurt the model performance on such datasets. We defer the results to the appendix. Diversity shift is considered as a different domain generalization problem with correlation shift in our paper, which is consistent with \cite{ye2022ood}. An example of diversity shift is that the images in the training domain are art paintings and cartoons, while they are photos in the test domain. We show in Section \ref{foundation} that ERM suffers from more severe performance degradation on the correlation-shifted datasets than on the diversity-shifted datasets in the o.o.d. case, and they may require different solutions. It is to the former that our paper proposes an effective solution.

We would also like to emphasize that the type of domain shift is independent of whether a dataset is real-world or not. Both \emph{CelebA-NS} and \emph{CelebA-HB} datasets in our paper are real-world datasets with correlation shifts. The correlation shift is more likely to arise from the selection bias in real-world scenarios, as simulated by the \emph{CelebA} datasets. However, since the datasets used in research are often shuffled, researchers tend to ignore the presence of selection bias (correlation shift), even though this is an important factor.

\subsection{Results.} Table \ref{table_corr} shows the performance of our approach under correlation shift. Under the DomainBed protocol, both ERM and the domain generalization algorithms officially reported by DomainBed do not perform well for correlation shift because they all suffer from a sharp performance drop when the test environment has reversed correlation with the training environment. Their accuracy rates are all very close, which is consistent with the results of \emph{CMNIST} reported in DomainBed \cite{gulrajani2020search}. For \emph{Colored MNIST}, on which the information-theoretic best accuracy is ${75\%}$ due to the ${25\%}$ noise, the accuracy of ERM and other domain generalization algorithms are no more than ${10.5\%}$ in the most difficult ``${-90\%}$" environment, which is far lower than random guess. In contrast, our DRM approach achieves ${69.7\%}$ accuracy, almost $60\%$ higher than the other algorithms. At the same time, our approach does not hurt performance on the ``${+90\%}$" and ``${+80\%}$" domains. On average, our approach outperforms ERM by $20\%$ and outperforms the best previous approach by $15\%$. For the other datasets, the results show the same pattern. Our approach is substantially ahead of the other methods by about $50\%$ on the most difficult domain and brings significant performance improvement of more than $15\%$ for the correlation shift problems. 

\begin{figure}[ht]
	\centering  
	\subcaptionbox{DSprites}{
		\includegraphics[width=0.31\linewidth]{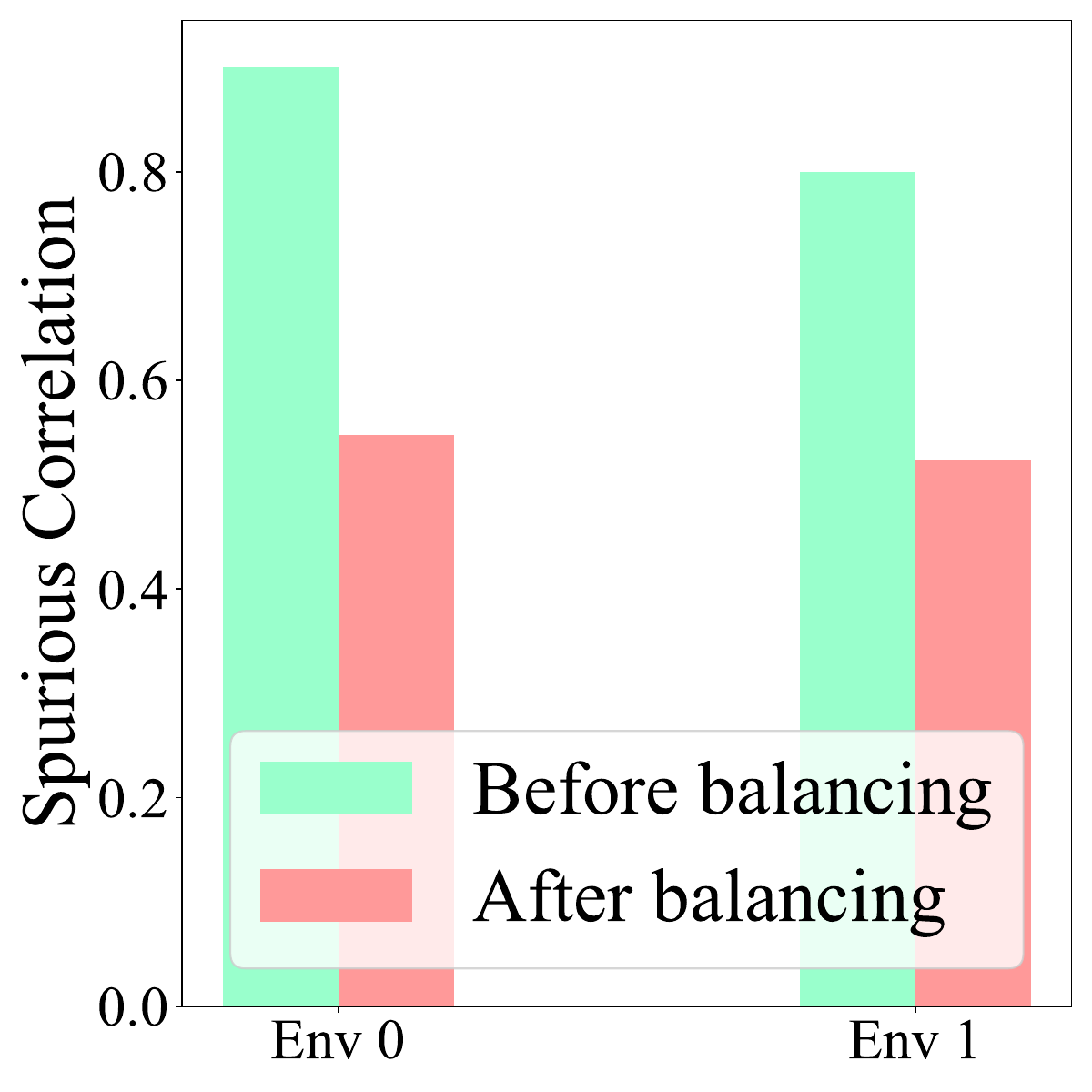}}
	\subcaptionbox{CelebA-HB}{
		\includegraphics[width=0.31\linewidth]{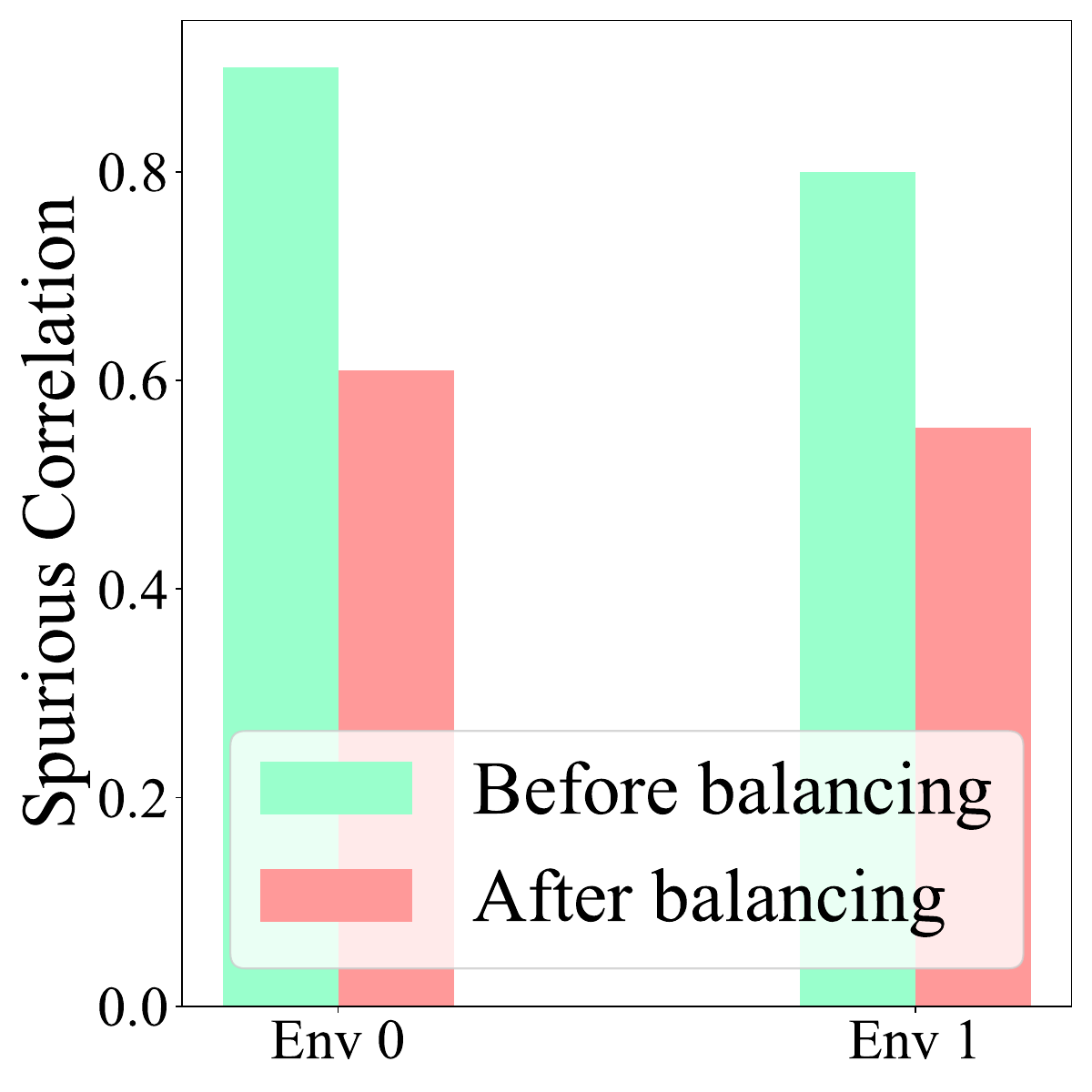}}
	\subcaptionbox{CelebA-NS}{
		\includegraphics[width=0.31\linewidth]{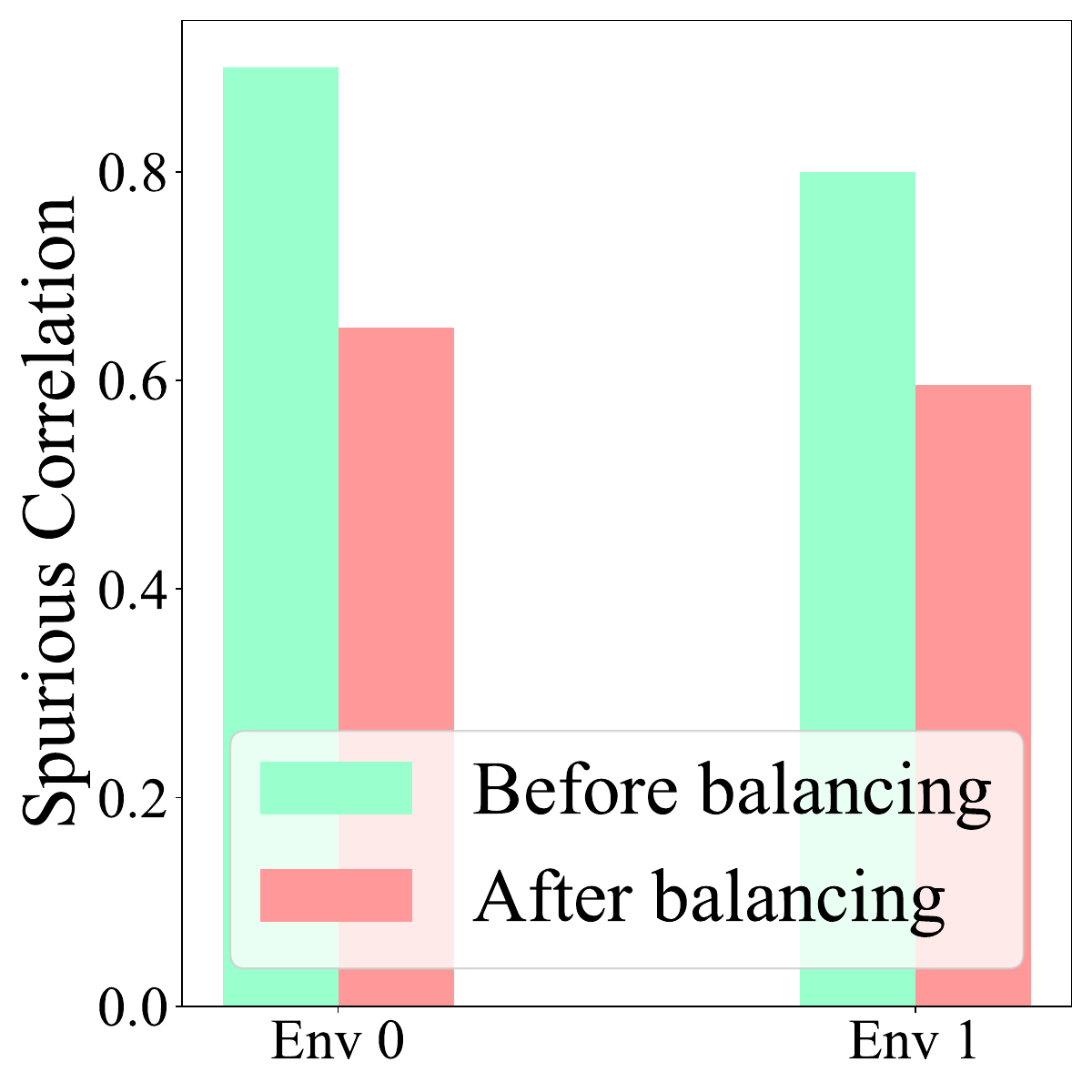}}

	\caption{The spurious correlation before and after balancing. }
	\label{corr}

\end{figure}

\begin{figure}[ht]
	\centering  
        {\includegraphics[width=0.11\linewidth]{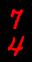}}
	{\includegraphics[width=0.11\linewidth]{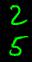}}
        {\includegraphics[width=0.11\linewidth]{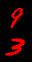}}
	{\includegraphics[width=0.11\linewidth]{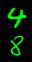}}
        {\includegraphics[width=0.11\linewidth]{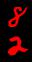}}
        {\includegraphics[width=0.11\linewidth]{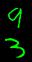}}
        {\includegraphics[width=0.11\linewidth]{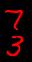}}
        {\includegraphics[width=0.11\linewidth]{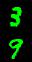}}
	\vspace{-0.1cm}
	\caption{Eight examples of balanced \emph{CMNIST} image pairs.}
	\label{cmist_sample}
\end{figure}

{\textbf{Results interpretation.} We attribute the significant performance improvement of DRM to its ability to reduce the spurious correlations in the training and validation set to a large extent, even on more realistic datasets, as shown in Figure \ref{corr}. We show eight examples of balanced \emph{CMNIST} image pairs in Figure \ref{cmist_sample}. The label-1 image with a red background is matched with the label-0 image with a red background, and the label-0 image with a green background is matched with the label-1 image with a green background, thus the correlation between the background color and the label is reduced. Models trained with balanced data are less susceptible to misleading correlations and so have better \ood generalization capabilities.}

\begin{table*}[t]
  \centering
  \caption{Experimental results on the correlation-shifted datasets, where the experiments are run by following the DomainBed setting. {\textbf{Min} and \textbf{Avg} are the minimum value and the average accuracy for all test environments, respectively.}} 
  \resizebox{1.0\textwidth}{!}{
    \begin{tabular}{lcccccccccc}
    \toprule
    \multicolumn{1}{c}{\multirow{2}[1]{*}{\textbf{Algorithm}}} & \multicolumn{2}{c}{\textbf{CMNIST}} & \multicolumn{2}{c}{\textbf{3DShapes}} & \multicolumn{2}{c}{\textbf{DSprites}} & \multicolumn{2}{c}{\textbf{CelebA-HB}} & \multicolumn{2}{c}{\textbf{CelebA-NS}} \\
\cmidrule{2-11}          & Min   & Avg   & Min   & Avg   & Min   & Avg   & Min   & Avg   & Min   & Avg \\
    \midrule
    ERM   & 10.0 ± 0.1 & 51.5 ± 0.1 & 10.1 ± 0.1 & 53.3 ± 0.1 & 13.8 ± 0.5 & 54.0 ± 0.1 & 16.8 ± 1.2 & 52.0 ± 0.5 & \multicolumn{1}{l}{21.1 ± 0.4} & 52.7 ± 0.5 \\
    IRM   & 10.2 ± 0.3 & 52.0 ± 0.1 & 10.0 ± 0.0 & 53.2 ± 0.1 & 14.5 ± 0.3 & 54.0 ± 0.1 & 20.4 ± 2.1 & 52.1 ± 0.7 & \multicolumn{1}{l}{21.5 ± 0.9} & 53.2 ± 0.4 \\
    GroupDRO & 10.0 ± 0.2 & 52.1 ± 0.0 & 10.5 ± 0.4 & 53.4 ± 0.1 & 15.0 ± 0.4 & 54.4 ± 0.2 & 18.3 ± 1.5 & 52.8 ± 0.9 & \multicolumn{1}{l}{21.2 ± 0.2} & 53.3 ± 0.2 \\
    Mixup & 10.1 ± 0.1 & 52.1 ± 0.2 & 10.2 ± 0.1 & 53.4 ± 0.2 & 14.0 ± 0.3 & 53.9 ± 0.0 & 17.9 ± 3.4 & 52.4 ± 1.1 & \multicolumn{1}{l}{22.2 ± 1.5} & 53.7 ± 0.7 \\
    MLDG  & 9.8 ± 0.1 & 51.5 ± 0.1 & 10.1 ± 0.1 & 53.5 ± 0.1 & 14.3 ± 0.3 & 54.2 ± 0.1 & 20.0 ± 2.1 & 53.0 ± 0.7 & \multicolumn{1}{l}{22.7 ± 1.7} & 53.7 ± 0.6 \\
    CORAL & 9.9 ± 0.1 & 51.5 ± 0.1 & 10.0 ± 0.0  & 53.3 ± 0.1 & 13.8 ± 0.2 & 53.9 ± 0.2 & 17.7 ± 1.6 & 52.4 ± 0.6 & \multicolumn{1}{l}{22.1 ± 1.1} & 53.4 ± 0.4 \\
    MMD   & 9.9 ± 0.3 & 51.5 ± 0.2 & 10.0 ± 0.1 & 53.2 ± 0.1 & 14.4 ± 0.0 & 51.4 ± 2.1 & 17.4 ± 1.8 & 50.7 ± 0.5 & \multicolumn{1}{l}{22.5 ± 0.6} & 53.3 ± 0.1 \\
    DANN  & 10.0 ± 0.0 & 51.5 ± 0.3 & 10.0 ± 0.0 & 53.3 ± 0.0 & 14.7 ± 0.3 & 54.1 ± 0.3 & 16.9 ± 1.7 & 51.7 ± 0.3 & \multicolumn{1}{l}{21.8 ± 1.5} & 53.7 ± 0.8 \\
    CDANN & 10.2 ± 0.1 & 51.7 ± 0.1 & 10.0 ± 0.0 & 53.3 ± 0.1 & 14.4 ± 0.2 & 54.0 ± 0.1 & 18.6 ± 2.6 & 52.5 ± 0.6 & \multicolumn{1}{l}{22.5 ± 1.2} & 53.9 ± 0.4 \\
    MTL   & 10.5 ± 0.1 & 51.4 ± 0.1 & 10.1 ± 0.0 & 53.4 ± 0.1 & 14.8 ± 0.5 & 54.3 ± 0.1 & 23.5 ± 1.4 & 53.7 ± 0.6 & \multicolumn{1}{l}{27.6 ± 1.2} & 54.9 ± 0.3 \\
    SagNet & 10.3 ± 0.1 & 51.7 ± 0.0 & 10.1 ± 0.1 & 53.4 ± 0.1 & 13.6 ± 0.1 & 54.0 ± 0.0 & 14.9 ± 0.9 & 50.4 ± 0.3 & \multicolumn{1}{l}{22.0 ± 0.6} & 53.1 ± 0.2 \\
    ARM   & 10.2 ± 0.0 & 56.2 ± 0.2 & 10.0 ± 0.0 & 55.2 ± 0.3 & 14.5 ± 0.6 & 59.7 ± 0.4 & 22.8 ± 2.3 & 54.1 ± 0.6 & \multicolumn{1}{l}{21.1 ± 1.4} & 53.0 ± 0.5 \\
    VREx  & 10.2 ± 0.0 & 51.8 ± 0.1 & 10.8 ± 0.3  & 53.5 ± 0.1 & 13.8 ± 0.3 & 53.9 ± 0.1 & 19.2 ± 1.9 & 52.5 ± 0.7 & \multicolumn{1}{l}{20.3 ± 0.4} & 53.2 ± 0.3 \\
    RSC   & 10.0 ± 0.2 & 51.7 ± 0.2 & 10.1 ± 0.1 & 53.2 ± 0.1 & 13.3 ± 0.2 & 53.8 ± 0.1 & 18.9 ± 1.1 & 52.5 ± 0.5 & \multicolumn{1}{l}{23.7 ± 0.8} & 54.3 ± 0.5 \\
    \midrule
    DRM (ours) & \textbf{69.7} ± 1.5 & \textbf{71.2} ± 0.6 & \textbf{74.5} ± 0.2 & \textbf{74.8} ± 0.1 & \textbf{73.3} ± 0.5 & \textbf{73.8} ± 0.2 & \textbf{61.0} ± 4.9 & \textbf{66.1} ± 0.6 & \textbf{59.9} ± 2.6 & \textbf{65.4} ± 1.2 \\
    \bottomrule
    \end{tabular}%
    }
  \label{table_corr}
\end{table*}%

\textbf{Ablation study.}
\label{ablation}
Our approach is based only on an improvement of the sampling phase, so both training set balancing (TB) and validation set balancing (VB) can be easily combined with other algorithms. In this section, we analyze the contribution of VB and TB in our approach, respectively. Results are shown in Table \ref{ab_cmnist}. Both training set balancing and validation set balancing can improve the \ood performance of the original model significantly, showing that both of them are effective and important components of our approach and can be used as a general framework to mitigate the correlation shift problem. VB demonstrates the ability to exploit the potential of existing methods such as IRM, as they can significantly outperform ERM simply by adding VB. When both VB and TB are used, the performances of existing methods achieve further improvements. ERM performs well in this case because VB and TB have largely eliminated spurious correlations.

\begin{table*}[htbp]
  \centering
  \caption{Ablation study for \emph{CMNIST}, \emph{DSprites}, and \emph{CelebA-HB}. VB stands for balancing during validation, while TB stands for balancing during training.}
  \resizebox{1.0\textwidth}{!}{
    \begin{tabular}{lcccccccccccc}
    \toprule
   \multicolumn{1}{c}{\multirow{2}[4]{*}{\textbf{Algorithm}}} & \multicolumn{4}{c}{\textbf{CMNIST}}    & \multicolumn{4}{c}{\textbf{DSprites}} & \multicolumn{4}{c}{\textbf{CelebA-HB}}\\

\cmidrule{2-13}          & +90\% & +80\% & -90\% & Avg   & +90\% & +80\% & -90\% & Avg & +90\% & +80\% & -90\% & Avg\\
    \midrule
    ERM   & 71.7 ± 0.1 & 72.9 ± 0.2 & 10.0 ± 0.1 & 51.5  & 73.5 ± 0.2    & 74.8 ± 0.0    & 13.8 ± 0.5    & 54.0 & 67.3 ± 0.2 & 71.8 ± 0.3 & 16.8 ± 1.2 & 52.0 \\
    ERM+VB & 71.5 ± 0.5 & 72.5 ± 0.3 & 30.7 ± 1.1 & 58.2 & 73.8 ± 0.2       & 74.4 ± 0.2       & 35.1 ± 2.2       & 61.1 & 68.3 ± 0.6 & 71.6 ± 0.8 & 37.4 ± 1.6 & 59.1 \\
    ERM+VB+TB & 71.4 ± 0.3 & 72.4 ± 0.4 & \textbf{69.7} ± 1.5 & \textbf{71.2}  & 73.7 ± 0.2    & 74.4 ± 0.1    & \textbf{73.3} ± 0.5    & \textbf{73.8} & 68.1 ± 1.0 & 69.3 ± 1.4 & \textbf{61.0} ± 4.9 & \textbf{66.1} \\
    \midrule
    IRM   &  72.5 ± 0.1 & 73.3 ± 0.5 & 10.2 ± 0.3 & 52.0 & 73.5 ± 0.2    & 74.2 ± 0.1    & 14.5 ± 0.3    & 54.0 & 65.9 ± 0.7 & 70.0 ± 0.7 & 20.4 ± 2.1 & 52.1 \\
    IRM+VB & 71.1 ± 0.6 & 72.9 ± 0.2 & 40.1 ± 5.2 & 61.4 & 73.9 ± 0.2       & 74.2 ± 0.1       & 50.1 ± 7.7       & 66.1 & 66.0 ± 0.9 & 71.1 ± 0.6 & 47.4 ± 4.1 & 61.5 \\
    IRM+VB+TB & 69.2 ± 0.6 & 71.7 ± 0.5 & \textbf{67.6} ± 1.5 & \textbf{69.5} & 73.9 ± 0.2       & 73.9 ± 0.2       & \textbf{71.3} ± 1.0       & \textbf{73.0}  & 66.0 ± 0.5 & 69.0 ± 1.5 & \textbf{65.7} ± 1.6 & \textbf{66.9} \\
    \midrule
    VREx  & 72.4 ± 0.3 & 72.9 ± 0.4 & 10.2 ± 0.0 & 51.8 & 73.4 ± 0.2    & 74.6 ± 0.1    & 13.8 ± 0.3    & 53.9 & 66.9 ± 0.3 & 71.4 ± 0.2 & 19.2 ± 1.8 & 52.5 \\
    VREx+VB & 72.5 ± 0.4 & 72.9 ± 0.1 & 49.9 ± 5.0 & 65.1 & 73.8 ± 0.1       & 74.4 ± 0.1       & 52.3 ± 7.9       & 66.8 & 67.4 ± 1.4 & 70.1 ± 1.1 & 46.8 ± 2.3 & 61.4 \\
    VREx+VB+TB & 71.2 ± 0.8 & 72.3 ± 0.2 & \textbf{68.3} ± 0.3 & \textbf{70.6}  & 73.7 ± 0.2       & 74.0 ± 0.1       & \textbf{72.9} ± 0.4       & \textbf{73.6} &  67.3 ± 0.8  &  68.3 ± 0.7 & \textbf{61.8} ± 3.0  & \textbf{65.8} \\
    \midrule
    CORAL & 71.6 ± 0.3 & 73.1 ± 0.1 & 9.9 ± 0.1 & 51.5 & 73.4 ± 0.3    & 74.5 ± 0.1    & 13.8 ± 0.2    & 53.9 & 68.3 ± 0.5 & 71.2 ± 0.2 & 17.7 ± 1.6 & 52.4 \\
    CORAL+VB & 71.4 ± 0.1 & 72.9 ± 0.3 & 33.2 ± 0.7 & 59.2 & 73.5 ± 0.1       & 74.5 ± 0.2       & 28.8 ± 2.2       & 59.0 & 65.9 ± 1.0 & 71.7 ± 1.1  & 47.3 ± 1.9 & 61.6  \\
    CORAL+VB+TB & 71.2 ± 0.8 & 72.8 ± 0.2 & \textbf{68.0} ± 0.8 & \textbf{70.7} & 73.8 ± 0.1       & 74.3 ± 0.2       & \textbf{71.9} ± 1.0       & \textbf{73.3} & 65.5 ± 0.9 & 68.8 ± 1.6 & \textbf{60.2} ± 4.6 & \textbf{64.8} \\
    \midrule
    CDANN & 72.0 ± 0.2  & 73.0 ± 0.2  & 10.2 ± 0.1 & 51.7 & 73.7 ± 0.3    & 74.0 ± 0.1    & 14.4 ± 0.2    & 54.0 & 66.8 ± 0.8 & 72.1 ± 0.4 & 18.6 ± 2.6    & 52.5 \\
    CDANN+VB & 72.6 ± 0.4 & 73.0 ± 0.1 & 45.9 ± 10.5 & 63.8 & 73.8 ± 0.2       & 74.5 ± 0.3       & 48.7 ± 6.8       & 65.7 & 66.3 ± 1.5 & 71.7 ± 0.7 & 48.4 ± 2.7 & 62.1  \\
    CDANN+VB+TB & 69.9 ± 0.2 & 71.1 ± 0.9  & \textbf{66.4} ± 0.7 & \textbf{69.1} & 74.1 ± 0.2       & 74.0 ± 0.1       & \textbf{73.0} ± 0.1       & \textbf{73.7} & 67.1 ± 0.5 & 70.3 ± 0.5 & \textbf{59.4} ± 3.1 & \textbf{65.6} \\
    \bottomrule
    \end{tabular}}
  \label{ab_cmnist}
  \vspace{-1em}
\end{table*}%

\begin{figure}[ht]
	\centering  
	\subcaptionbox{IRM}{
		\includegraphics[width=1\linewidth]{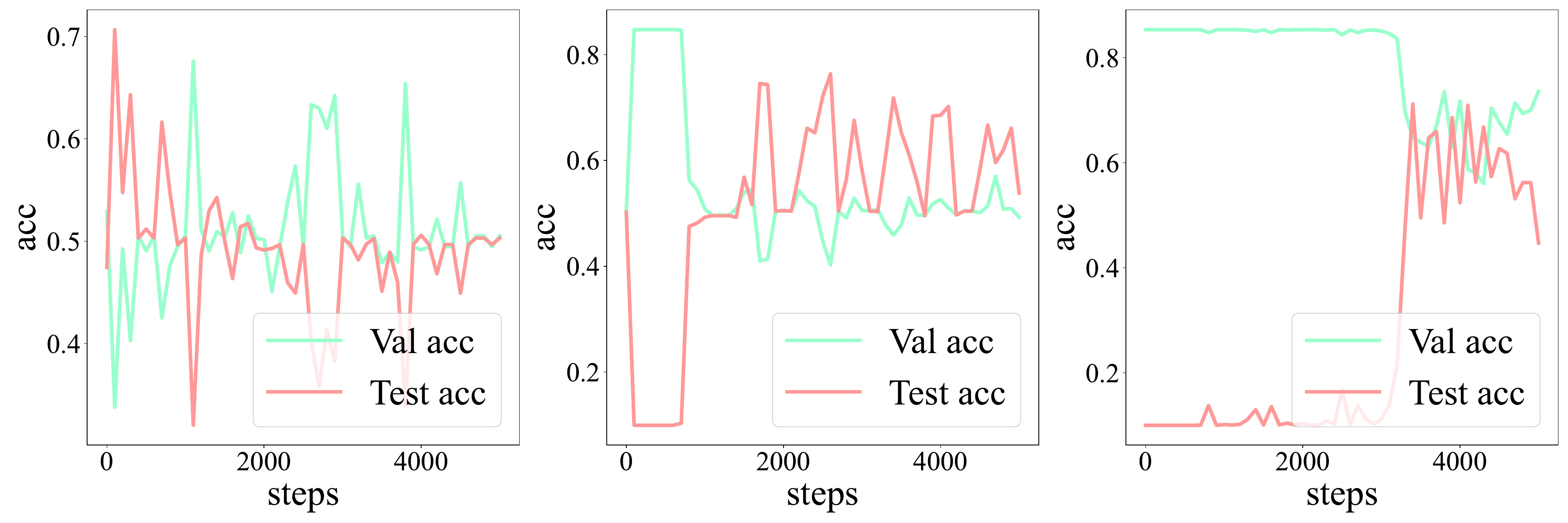}}

	\subcaptionbox{DRM}{
		\includegraphics[width=1\linewidth]{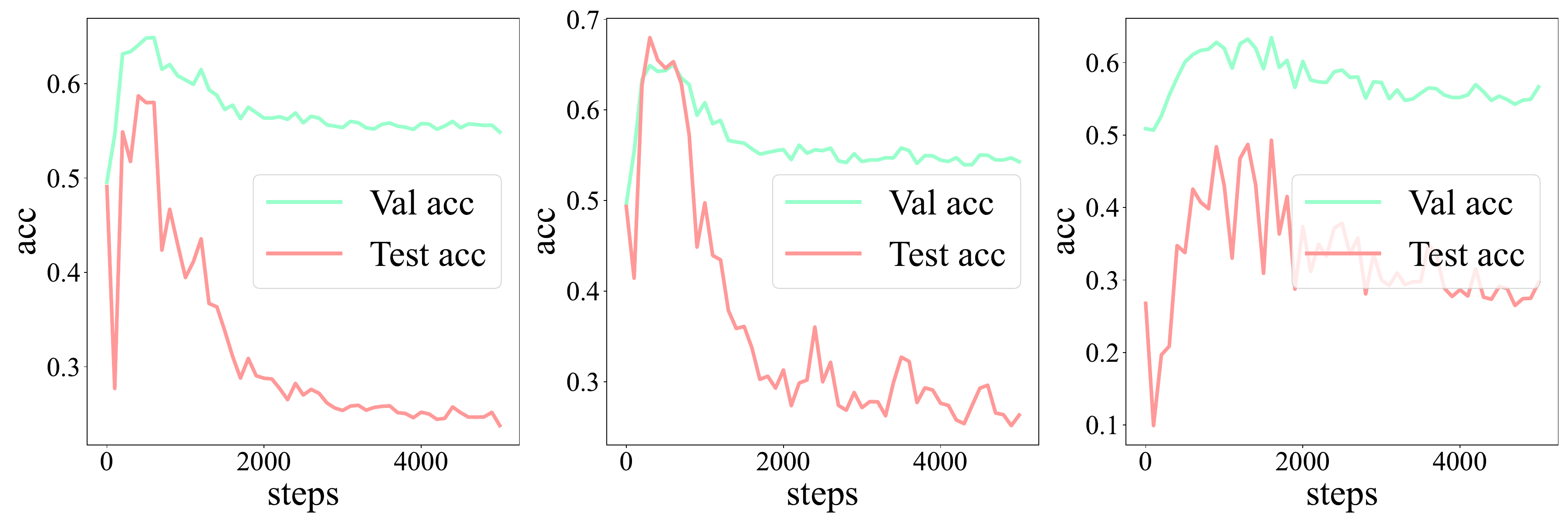}}

	\caption{Accuracy curves of IRM and our DRM with different random seed for hyperparameter search.}
	\label{DomainBedfig}

\end{figure}

{\textbf{Accuracy curves analysis.}
Many mainstream domain generalization methods design new loss functions by incorporating invariant constraints, such as IRM and VREx, which leads to a very unstable training process. In contrast, as shown in Figure \ref{DomainBedfig}, the fluctuation of the accuracy curve of our DRM in the training phase is relatively small. The validation accuracy is basically consistent with the test accuracy, showing the same trend in Figure \ref{DomainBedfig}.
}

\subsection{Foundation Models and \Ood Generalization}
\label{foundation}
The scale of data is very important for the domain generalization problem. In this section, we show that large-scale models pre-trained with large-scale data can generalize well on diversity-shifted datasets, but show little improvement on correlation-shifted datasets. We choose the \emph{Colored MNIST} dataset and the \emph{CelebA-HB} dataset for correlation shift, and the \emph{PACS} dataset and the \emph{Office-Home} dataset for diversity shift. We conduct experiments on the large-scale pre-trained model CLIP \cite{radford2021learning} and EVA \cite{fang2022eva} by linear probing. As shown in Table \ref{Foundation model}, CLIP (ViT-L\cite{dosovitskiy2020image}) and EVA (ViT, 1B parameters) in the \ood case even outperforms ResNet-50 in the \iid case on \emph{PACS} and \emph{Office-Home}, suggesting that the diversity shift problem can be alleviated by increasing the scale of training data and parameters, as well as improving the model architecture. However, they cannot generalize well in the correlation shift case. Therefore, we believe that our approach closes the gap between foundation models and domain generalization to some extent.

\begin{table}[htbp]
  \centering
  \caption{Results of Different Methods on Typical Correlation and Diversity-shifted Datasets.}
  \vspace{-0.1cm}
  \resizebox{1.0\linewidth}{!}{
    \begin{tabular}{clcccc}
    \toprule
    \multirow{2}[4]{*}{\textbf{\ood}} & \multicolumn{1}{c}{\multirow{2}[4]{*}{\textbf{Method}}} & \multicolumn{2}{c}{\textbf{Correlation Shift}} & \multicolumn{2}{c}{\textbf{Diversity Shift}} \\
\cmidrule{3-6}          &       & \textbf{CMNIST} & \textbf{CelebA-HB} & \textbf{PACS}  & \textbf{Office-Home} \\
    \midrule
    \XSolidBrush & ERM (RN50) & 86.6  & 83.6  & 96.3  & 80.4 \\
    \Checkmark & ERM (RN50) & 51.5 (-35.1) & 52.0 (-31.6) & 85.5 (-10.8) & 66.5 (-13.9) \\
    \Checkmark & CLIP (RN50) & 48.9 (-37.7) & 55.8 (-27.8) & 64.1 (-32.2) & 48.3 (-32.1) \\
    \Checkmark & CLIP (ViT-B) & 52.2 (-34.4) & 54.8 (-28.8) & {95.7} (-0.6) & {82.3} (+1.9) \\
    \Checkmark & CLIP (ViT-L) & 52.5 (-34.1) & 53.8 (-29.8) & {98.4} (+2.1) & {88.3} (+7.9) \\
    \Checkmark & EVA (ViT, 1B Paras) & 51.3 (-35.3) & 53.6 (-30.0) & \textbf{98.7} (+2.4) & \textbf{90.7} (+10.3) \\
    \Checkmark & Our DRM (RN50)  & \textbf{71.2} (-15.4) & \textbf{66.1} (-17.5) & 84.8 (-11.5) & 65.7 (-14.7) \\
    \bottomrule
    \end{tabular}}
  \label{Foundation model}
\end{table}

\begin{figure}[ht]
	\centering  
	\subcaptionbox{CelebA-HB}[.49\linewidth]{
		\includegraphics[width=1\linewidth]{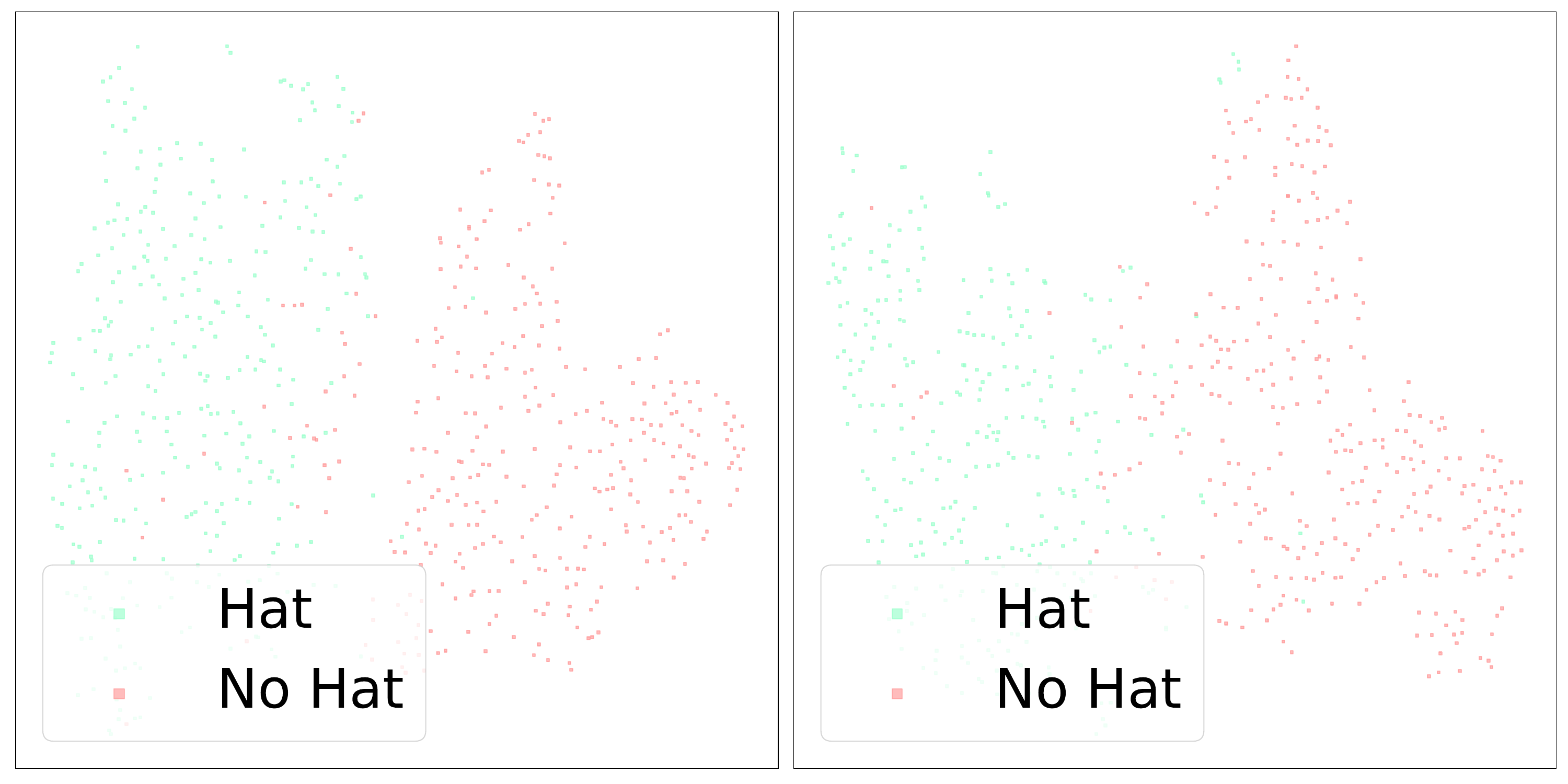}}
	\subcaptionbox{CelebA-NS}[.49\linewidth]{
		\includegraphics[width=1\linewidth]{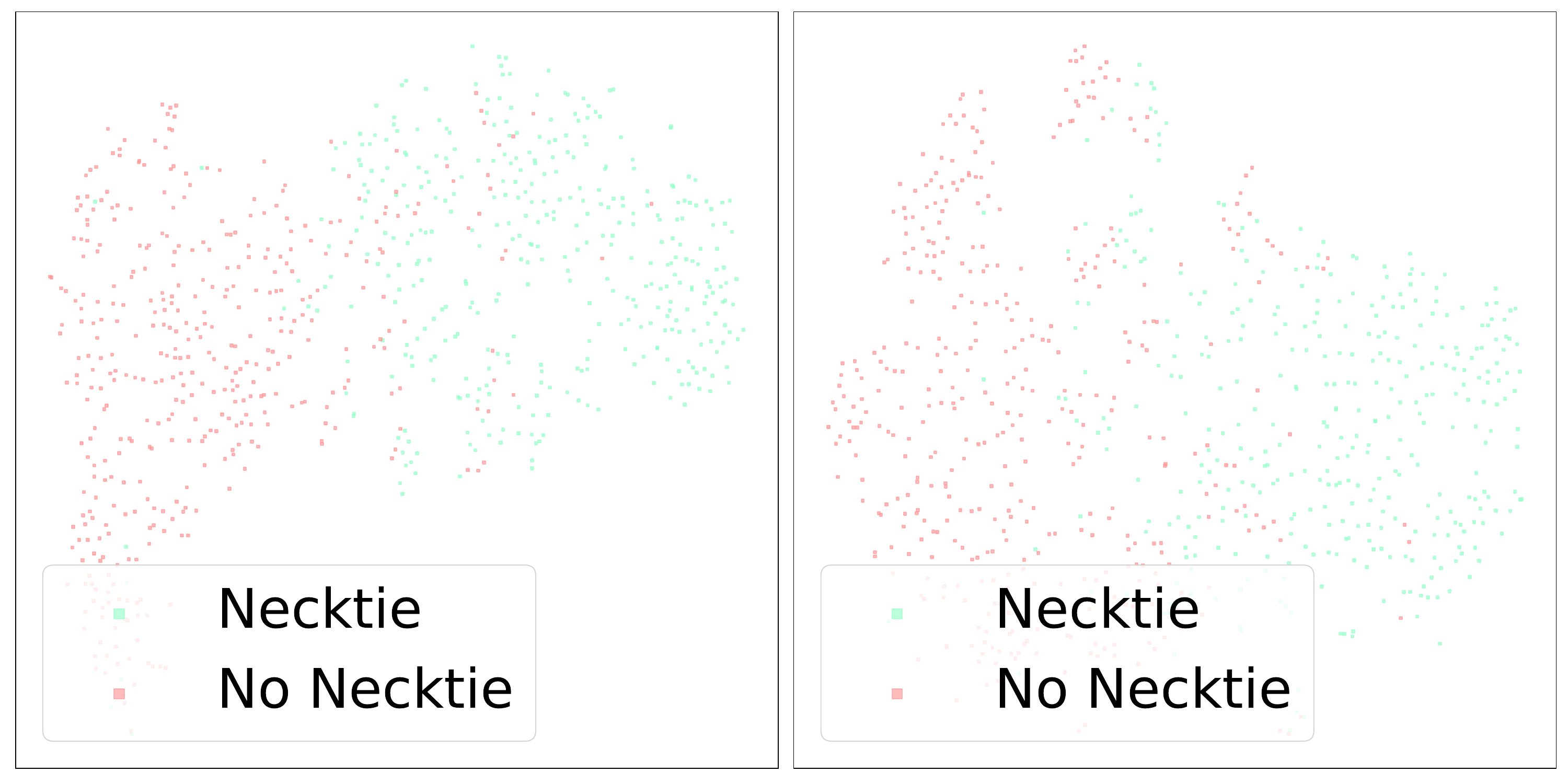}}
	\vspace{-0.1cm}
	\caption{Visualization of 2-D t-SNE result of $Z$. Different colors represent different spurious features, ``Wearing Hat'' vs. ``No Hat'', or ``Wearing Neckline'' vs. ``No Neckline''. {Each subfigure represents the results of two training domains when ``${-90\%}$'' is the test domain, which is the most difficult for the model to generalize.} }
	\label{scatter_diagram}
\end{figure}

\subsection{Visual Explanation}
In this section, we choose the \emph{CelebA-HB} dataset and the \emph{CelebA-NS} dataset to visualize the results. For \emph{CelebA-HB}, the indirect effect is the pathway between the spurious feature and label, which is ``Wearing Hat'' and ``No Beard'', respectively. For \emph{CelebA-NS}, the indirect effect is the pathway between ``Wearing Necktie'' and ``Smiling''.


\textbf{Analysis of indirect effect representation. }
Our DRM approach recovers the indirect effect in the first stage. Thus the quality of indirect effect representation $\hat{Z}$ is important. We use t-SNE \cite{van2008visualizing} to reduce the dimension of $\hat{Z}$ extracted in the first stage to 2 and show the results in Figure \ref{scatter_diagram}. We can observe that data points with the same spurious feature show a clustering effect, which means that $\hat{Z}$ is an appropriate representation of the spurious feature.

\begin{figure}[htbp]
    \centering
    \includegraphics[width=1\linewidth]{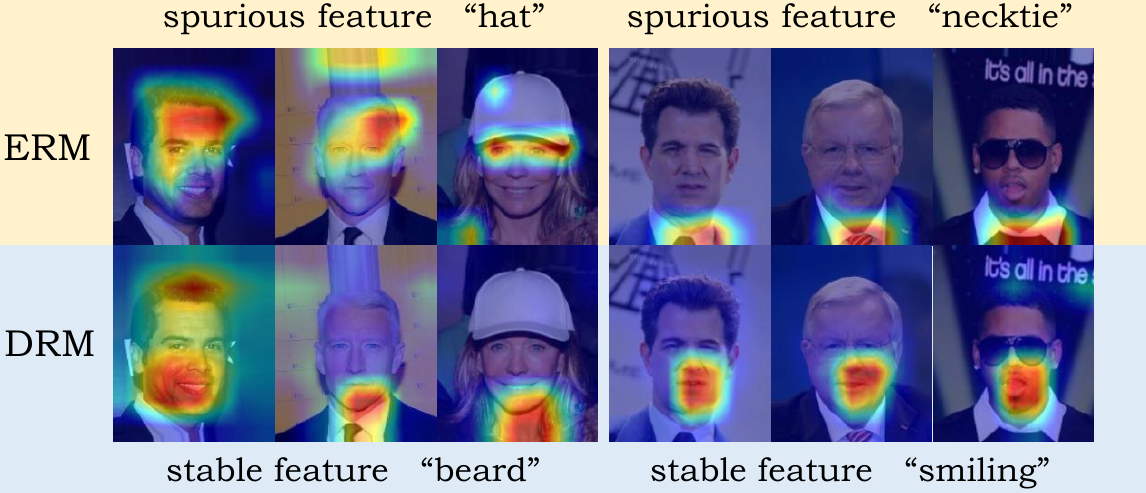}
    \caption{Attention maps of ERM and our method, respectively. \textbf{The left half} is on the \emph{CelebA-HB} dataset. \textbf{The right half} is on the \emph{CelebA-NS} dataset.}
    \label{attention_map}
\end{figure}

\textbf{Attention map. } 
In Figure \ref{attention_map}, we present the attention maps of the last convolution layer for ERM (the first row) and DRM (the second row). The model trained by ERM focuses on the spurious feature ``Wearing Hat'' and ``Wearing Necktie'', while the model trained by DRM focuses on the stable feature ``No Beard'' and ``Smiling''.

\section{Related Works}
\textbf{Domain generalization with causality.} A large body of works has introduced tools from causality inference to the domain generalization problem. Causality has been shown to be robust across domains \cite{peters2016causal}, and some works discussed which causal factors can be extracted \cite{scholkopf2012causal,scholkopf2021toward} and the connection between causality and generalization \cite{https://doi.org/10.48550/arxiv.2006.07433}. \cite{pearl2001direct} gave a definition of natural direct effects. Although the concept has been introduced in early works \cite{zhang2018fairness,qi2020two,heskes2020causal}, no work analyzed domain generalization using this framework.

\textbf{Matching based methods.} Matching is a common approach that aims to eliminate selection bias in causal inference by matching comparable instances \cite{rosenbaum1983central}. \cite{mahajan2021domain} proposed an unsupervised matching algorithm and \cite{wang2022causal} introduced the propensity score matching method to balance the mini-batch. Our method also uses a mini-batch balancing approach in the second stage. However, we extract the indirect-effect representation in the first stage by learning a domain discriminator and propose a balancing-based model selection method, which are different from above approaches and help our approach to achieve better performance.

\section{Conclusion}
In this paper, we introduce the concept of direct and indirect effects from causal inference to the domain generalization. We propose a domain generalization method to extract the indirect-effect representation and remove the indirect effects during training. {We also propose a new approach to do model selection in the \ood setting. Both our domain generalization approach and model selection approach can be combined with other existing algorithms and improve their performance significantly.} Experimental results show that our approach achieves the state-of-the-art performance.


{\small
\bibliographystyle{ieee_fullname}
\bibliography{Paper}

\begin{thebibliography}{100}\itemsep=-1pt

\bibitem{ahuja2021invariance}
Kartik Ahuja, Ethan Caballero, Dinghuai Zhang, Jean-Christophe Gagnon-Audet,
  Yoshua Bengio, Ioannis Mitliagkas, and Irina Rish.
\newblock Invariance principle meets information bottleneck for
  out-of-distribution generalization.
\newblock {\em Advances in Neural Information Processing Systems},
  34:3438--3450, 2021.

\bibitem{ahuja2020irmg}
Kartik Ahuja, Karthikeyan Shanmugam, Kush Varshney, and Amit Dhurandhar.
\newblock Invariant risk minimization games.
\newblock In {\em International Conference on Machine Learning}, pages
  145--155. PMLR, 2020.

\bibitem{ahuja2020empirical}
Kartik Ahuja, Jun Wang, Amit Dhurandhar, Karthikeyan Shanmugam, and Kush~R
  Varshney.
\newblock Empirical or invariant risk minimization? a sample complexity
  perspective.
\newblock In {\em International Conference on Learning Representations}, 2020.

\bibitem{albuquerque2019generalizing}
Isabela Albuquerque, Jo{\~a}o Monteiro, Mohammad Darvishi, Tiago~H Falk, and
  Ioannis Mitliagkas.
\newblock Generalizing to unseen domains via distribution matching.
\newblock {\em arXiv preprint arXiv:1911.00804}, 2019.

\bibitem{Alpher02}
Alvin Alpher.
\newblock Frobnication.
\newblock {\em Journal of Foo}, 12(1):234--778, 2002.

\bibitem{Alpher03}
Alvin Alpher and Ferris P.~N. Fotheringham-Smythe.
\newblock Frobnication revisited.
\newblock {\em Journal of Foo}, 13(1):234--778, 2003.

\bibitem{Alpher04}
Alvin Alpher, Ferris P.~N. Fotheringham-Smythe, and Gavin Gamow.
\newblock Can a machine frobnicate?
\newblock {\em Journal of Foo}, 14(1):234--778, 2004.

\bibitem{arjovsky2019invariant}
Martin Arjovsky, L{\'e}on Bottou, Ishaan Gulrajani, and David Lopez-Paz.
\newblock Invariant risk minimization.
\newblock {\em arXiv preprint arXiv:1907.02893}, 2019.

\bibitem{balaji2018metareg}
Yogesh Balaji, Swami Sankaranarayanan, and Rama Chellappa.
\newblock Metareg: Towards domain generalization using meta-regularization.
\newblock {\em Advances in neural information processing systems}, 31, 2018.

\bibitem{beery2018recognition}
Sara Beery, Grant~Van Horn, and Pietro Perona.
\newblock Recognition in terra incognita.
\newblock In {\em European Conference on Computer Vision}, pages 472--489.
  Springer, 2018.

\bibitem{bellot2020accounting}
Alexis Bellot and Mihaela van~der Schaar.
\newblock Accounting for unobserved confounding in domain generalization.
\newblock {\em arXiv preprint arXiv:2007.10653}, 2020.

\bibitem{ben2006analysis}
Shai Ben-David, John Blitzer, Koby Crammer, and Fernando Pereira.
\newblock Analysis of representations for domain adaptation.
\newblock {\em Advances in neural information processing systems}, 19, 2006.

\bibitem{blanchard2011generalizing}
Gilles Blanchard, Gyemin Lee, and Clayton Scott.
\newblock Generalizing from several related classification tasks to a new
  unlabeled sample.
\newblock {\em Advances in neural information processing systems}, 24, 2011.

\bibitem{bui2021exploiting}
Manh-Ha Bui, Toan Tran, Anh Tran, and Dinh Phung.
\newblock Exploiting domain-specific features to enhance domain generalization.
\newblock {\em Advances in Neural Information Processing Systems},
  34:21189--21201, 2021.

\bibitem{3dshapes18}
Chris Burgess and Hyunjik Kim.
\newblock 3d shapes dataset.
\newblock https://github.com/deepmind/3dshapes-dataset/, 2018.

\bibitem{carlucci2019domain}
Fabio~M Carlucci, Antonio D'Innocente, Silvia Bucci, Barbara Caputo, and
  Tatiana Tommasi.
\newblock Domain generalization by solving jigsaw puzzles.
\newblock In {\em Proceedings of the IEEE/CVF Conference on Computer Vision and
  Pattern Recognition}, pages 2229--2238, 2019.

\bibitem{cha2021swad}
Junbum Cha, Sanghyuk Chun, Kyungjae Lee, Han-Cheol Cho, Seunghyun Park, Yunsung
  Lee, and Sungrae Park.
\newblock Swad: Domain generalization by seeking flat minima.
\newblock {\em Advances in Neural Information Processing Systems},
  34:22405--22418, 2021.

\bibitem{chevalley2022invariant}
Mathieu Chevalley, Charlotte Bunne, Andreas Krause, and Stefan Bauer.
\newblock Invariant causal mechanisms through distribution matching.
\newblock {\em arXiv preprint arXiv:2206.11646}, 2022.

\bibitem{https://doi.org/10.48550/arxiv.2006.07433}
Rune Christiansen, Niklas Pfister, Martin~Emil Jakobsen, Nicola Gnecco, and
  Jonas Peters.
\newblock A causal framework for distribution generalization, 2020.

\bibitem{dosovitskiy2020image}
Alexey Dosovitskiy, Lucas Beyer, Alexander Kolesnikov, Dirk Weissenborn,
  Xiaohua Zhai, Thomas Unterthiner, Mostafa Dehghani, Matthias Minderer, Georg
  Heigold, Sylvain Gelly, et~al.
\newblock An image is worth 16x16 words: Transformers for image recognition at
  scale.
\newblock {\em arXiv preprint arXiv:2010.11929}, 2020.

\bibitem{dou2019domain}
Qi Dou, Daniel Coelho~de Castro, Konstantinos Kamnitsas, and Ben Glocker.
\newblock Domain generalization via model-agnostic learning of semantic
  features.
\newblock {\em Advances in Neural Information Processing Systems}, 32, 2019.

\bibitem{fang2013unbiased}
Chen Fang, Ye Xu, and Daniel~N Rockmore.
\newblock Unbiased metric learning: On the utilization of multiple datasets and
  web images for softening bias.
\newblock In {\em Proceedings of the IEEE International Conference on Computer
  Vision}, pages 1657--1664, 2013.

\bibitem{fang2022eva}
Yuxin Fang, Wen Wang, Binhui Xie, Quan Sun, Ledell Wu, Xinggang Wang, Tiejun
  Huang, Xinlong Wang, and Yue Cao.
\newblock Eva: Exploring the limits of masked visual representation learning at
  scale.
\newblock {\em arXiv preprint arXiv:2211.07636}, 2022.

\bibitem{ganin2016domain}
Yaroslav Ganin, Evgeniya Ustinova, Hana Ajakan, Pascal Germain, Hugo
  Larochelle, Fran{\c{c}}ois Laviolette, Mario Marchand, and Victor Lempitsky.
\newblock Domain-adversarial training of neural networks.
\newblock {\em The journal of machine learning research}, 17(1):2096--2030,
  2016.

\bibitem{ghifary2015domain}
Muhammad Ghifary, W~Bastiaan Kleijn, Mengjie Zhang, and David Balduzzi.
\newblock Domain generalization for object recognition with multi-task
  autoencoders.
\newblock In {\em Proceedings of the IEEE international conference on computer
  vision}, pages 2551--2559, 2015.

\bibitem{gulrajani2020search}
Ishaan Gulrajani and David Lopez-Paz.
\newblock In search of lost domain generalization.
\newblock In {\em International Conference on Learning Representations}, 2020.

\bibitem{he2016deep}
Kaiming He, Xiangyu Zhang, Shaoqing Ren, and Jian Sun.
\newblock Deep residual learning for image recognition. cvpr. 2016.
\newblock {\em arXiv preprint arXiv:1512.03385}, 2016.

\bibitem{heskes2020causal}
Tom Heskes, Evi Sijben, Ioan~Gabriel Bucur, and Tom Claassen.
\newblock Causal shapley values: Exploiting causal knowledge to explain
  individual predictions of complex models.
\newblock {\em Advances in neural information processing systems},
  33:4778--4789, 2020.

\bibitem{huang2020self}
Zeyi Huang, Haohan Wang, Eric~P Xing, and Dong Huang.
\newblock Self-challenging improves cross-domain generalization.
\newblock In {\em European Conference on Computer Vision}, pages 124--140.
  Springer, 2020.

\bibitem{iwasawa2021test}
Yusuke Iwasawa and Yutaka Matsuo.
\newblock Test-time classifier adjustment module for model-agnostic domain
  generalization.
\newblock {\em Advances in Neural Information Processing Systems},
  34:2427--2440, 2021.

\bibitem{kamath2021does}
Pritish Kamath, Akilesh Tangella, Danica Sutherland, and Nathan Srebro.
\newblock Does invariant risk minimization capture invariance?
\newblock In {\em International Conference on Artificial Intelligence and
  Statistics}, pages 4069--4077. PMLR, 2021.

\bibitem{kim2021selfreg}
Daehee Kim, Youngjun Yoo, Seunghyun Park, Jinkyu Kim, and Jaekoo Lee.
\newblock Selfreg: Self-supervised contrastive regularization for domain
  generalization.
\newblock In {\em Proceedings of the IEEE/CVF International Conference on
  Computer Vision}, pages 9619--9628, 2021.

\bibitem{krueger2021out}
David Krueger, Ethan Caballero, Joern-Henrik Jacobsen, Amy Zhang, Jonathan
  Binas, Dinghuai Zhang, Remi Le~Priol, and Aaron Courville.
\newblock Out-of-distribution generalization via risk extrapolation (rex).
\newblock In {\em International Conference on Machine Learning}, pages
  5815--5826. PMLR, 2021.

\bibitem{lecun1998mnist}
Yann LeCun.
\newblock The mnist database of handwritten digits.
\newblock {\em http://yann. lecun. com/exdb/mnist/}, 1998.

\bibitem{li2018learning}
Da Li, Yongxin Yang, Yi-Zhe Song, and Timothy Hospedales.
\newblock Learning to generalize: Meta-learning for domain generalization.
\newblock In {\em Proceedings of the AAAI conference on artificial
  intelligence}, volume~32, 2018.

\bibitem{li2017deeper}
Da Li, Yongxin Yang, Yi-Zhe Song, and Timothy~M Hospedales.
\newblock Deeper, broader and artier domain generalization.
\newblock In {\em Proceedings of the IEEE international conference on computer
  vision}, pages 5542--5550, 2017.

\bibitem{li2018domain}
Haoliang Li, Sinno~Jialin Pan, Shiqi Wang, and Alex~C Kot.
\newblock Domain generalization with adversarial feature learning.
\newblock In {\em Proceedings of the IEEE conference on computer vision and
  pattern recognition}, pages 5400--5409, 2018.

\bibitem{li2018mmd}
Haoliang Li, Sinno~Jialin Pan, Shiqi Wang, and Alex~C Kot.
\newblock Domain generalization with adversarial feature learning.
\newblock In {\em Proceedings of the IEEE conference on computer vision and
  pattern recognition}, pages 5400--5409, 2018.

\bibitem{li2020domain}
Haoliang Li, YuFei Wang, Renjie Wan, Shiqi Wang, Tie-Qiang Li, and Alex Kot.
\newblock Domain generalization for medical imaging classification with
  linear-dependency regularization.
\newblock {\em Advances in Neural Information Processing Systems},
  33:3118--3129, 2020.

\bibitem{li2018deep}
Ya Li, Xinmei Tian, Mingming Gong, Yajing Liu, Tongliang Liu, Kun Zhang, and
  Dacheng Tao.
\newblock Deep domain generalization via conditional invariant adversarial
  networks.
\newblock In {\em Proceedings of the European Conference on Computer Vision
  (ECCV)}, pages 624--639, 2018.

\bibitem{li2019feature}
Yiying Li, Yongxin Yang, Wei Zhou, and Timothy Hospedales.
\newblock Feature-critic networks for heterogeneous domain generalization.
\newblock In {\em International Conference on Machine Learning}, pages
  3915--3924. PMLR, 2019.

\bibitem{lin2022bayesian}
Yong Lin, Hanze Dong, Hao Wang, and Tong Zhang.
\newblock Bayesian invariant risk minimization.
\newblock In {\em Proceedings of the IEEE/CVF Conference on Computer Vision and
  Pattern Recognition}, pages 16021--16030, 2022.

\bibitem{liu2021learning}
Chang Liu, Xinwei Sun, Jindong Wang, Haoyue Tang, Tao Li, Tao Qin, Wei Chen,
  and Tie-Yan Liu.
\newblock Learning causal semantic representation for out-of-distribution
  prediction.
\newblock {\em Advances in Neural Information Processing Systems},
  34:6155--6170, 2021.

\bibitem{liu2015faceattributes}
Ziwei Liu, Ping Luo, Xiaogang Wang, and Xiaoou Tang.
\newblock Deep learning face attributes in the wild.
\newblock In {\em Proceedings of International Conference on Computer Vision
  (ICCV)}, December 2015.

\bibitem{lu2021invariant}
Chaochao Lu, Yuhuai Wu, Jos{\'e}~Miguel Hern{\'a}ndez-Lobato, and Bernhard
  Sch{\"o}lkopf.
\newblock Invariant causal representation learning for out-of-distribution
  generalization.
\newblock In {\em International Conference on Learning Representations}, 2021.

\bibitem{lv2022causality}
Fangrui Lv, Jian Liang, Shuang Li, Bin Zang, Chi~Harold Liu, Ziteng Wang, and
  Di Liu.
\newblock Causality inspired representation learning for domain generalization.
\newblock In {\em Proceedings of the IEEE/CVF Conference on Computer Vision and
  Pattern Recognition}, pages 8046--8056, 2022.

\bibitem{mahajan2021domain}
Divyat Mahajan, Shruti Tople, and Amit Sharma.
\newblock Domain generalization using causal matching.
\newblock In {\em International Conference on Machine Learning}, pages
  7313--7324. PMLR, 2021.

\bibitem{dsprites17}
Loic Matthey, Irina Higgins, Demis Hassabis, and Alexander Lerchner.
\newblock dsprites: Disentanglement testing sprites dataset.
\newblock https://github.com/deepmind/dsprites-dataset/, 2017.

\bibitem{motiian2017unified}
Saeid Motiian, Marco Piccirilli, Donald~A Adjeroh, and Gianfranco Doretto.
\newblock Unified deep supervised domain adaptation and generalization.
\newblock In {\em Proceedings of the IEEE international conference on computer
  vision}, pages 5715--5725, 2017.

\bibitem{muandet2013domain}
Krikamol Muandet, David Balduzzi, and Bernhard Sch{\"o}lkopf.
\newblock Domain generalization via invariant feature representation.
\newblock In {\em International Conference on Machine Learning}, pages 10--18.
  PMLR, 2013.

\bibitem{nagarajan2020understanding}
Vaishnavh Nagarajan, Anders Andreassen, and Behnam Neyshabur.
\newblock Understanding the failure modes of out-of-distribution
  generalization.
\newblock In {\em International Conference on Learning Representations}, 2020.

\bibitem{nam2019reducing}
Hyeonseob Nam, HyunJae Lee, Jongchan Park, Wonjun Yoon, and Donggeun Yoo.
\newblock Reducing domain gap via style-agnostic networks.
\newblock {\em arXiv preprint arXiv:1910.11645}, 2(7):8, 2019.

\bibitem{Authors14}
Full~Author Name.
\newblock The frobnicatable foo filter, 2014.
\newblock Face and Gesture submission ID 324. Supplied as additional material
  {\tt fg324.pdf}.

\bibitem{Authors14b}
Full~Author Name.
\newblock Frobnication tutorial, 2014.
\newblock Supplied as additional material {\tt tr.pdf}.

\bibitem{parascandolo2020learning}
Giambattista Parascandolo, Alexander Neitz, Antonio Orvieto, Luigi Gresele, and
  Bernhard Sch{\"o}lkopf.
\newblock Learning explanations that are hard to vary.
\newblock {\em arXiv preprint arXiv:2009.00329}, 2020.

\bibitem{pearl2001direct}
Judea Pearl.
\newblock Direct and indirect effects.
\newblock {\em Probabilistic and Causal Inference: The Works of Judea Pearl},
  page 373, 2001.

\bibitem{pearl2009causality}
Judea Pearl.
\newblock {\em Causality}.
\newblock Cambridge university press, 2009.

\bibitem{peng2019moment}
Xingchao Peng, Qinxun Bai, Xide Xia, Zijun Huang, Kate Saenko, and Bo Wang.
\newblock Moment matching for multi-source domain adaptation.
\newblock In {\em Proceedings of the IEEE/CVF international conference on
  computer vision}, pages 1406--1415, 2019.

\bibitem{peters2016causal}
Jonas Peters, Peter B{\"u}hlmann, and Nicolai Meinshausen.
\newblock Causal inference by using invariant prediction: identification and
  confidence intervals.
\newblock {\em Journal of the Royal Statistical Society: Series B (Statistical
  Methodology)}, 78(5):947--1012, 2016.

\bibitem{piratla2020efficient}
Vihari Piratla, Praneeth Netrapalli, and Sunita Sarawagi.
\newblock Efficient domain generalization via common-specific low-rank
  decomposition.
\newblock In {\em International Conference on Machine Learning}, pages
  7728--7738. PMLR, 2020.

\bibitem{qi2020two}
Jiaxin Qi, Yulei Niu, Jianqiang Huang, and Hanwang Zhang.
\newblock Two causal principles for improving visual dialog.
\newblock In {\em Proceedings of the IEEE/CVF conference on computer vision and
  pattern recognition}, pages 10860--10869, 2020.

\bibitem{qiao2020learning}
Fengchun Qiao, Long Zhao, and Xi Peng.
\newblock Learning to learn single domain generalization.
\newblock In {\em Proceedings of the IEEE/CVF Conference on Computer Vision and
  Pattern Recognition}, pages 12556--12565, 2020.

\bibitem{radford2021learning}
Alec Radford, Jong~Wook Kim, Chris Hallacy, Aditya Ramesh, Gabriel Goh,
  Sandhini Agarwal, Girish Sastry, Amanda Askell, Pamela Mishkin, Jack Clark,
  et~al.
\newblock Learning transferable visual models from natural language
  supervision.
\newblock In {\em International conference on machine learning}, pages
  8748--8763. PMLR, 2021.

\bibitem{DBLP:journals/corr/abs-2103-00020}
Alec Radford, Jong~Wook Kim, Chris Hallacy, Aditya Ramesh, Gabriel Goh,
  Sandhini Agarwal, Girish Sastry, Amanda Askell, Pamela Mishkin, Jack Clark,
  Gretchen Krueger, and Ilya Sutskever.
\newblock Learning transferable visual models from natural language
  supervision.
\newblock {\em CoRR}, abs/2103.00020, 2021.

\bibitem{rame2022fishr}
Alexandre Rame, Corentin Dancette, and Matthieu Cord.
\newblock Fishr: Invariant gradient variances for out-of-distribution
  generalization.
\newblock In {\em International Conference on Machine Learning}, pages
  18347--18377. PMLR, 2022.

\bibitem{robey2021model}
Alexander Robey, George~J Pappas, and Hamed Hassani.
\newblock Model-based domain generalization.
\newblock {\em Advances in Neural Information Processing Systems},
  34:20210--20229, 2021.

\bibitem{rosenbaum1983central}
Paul~R Rosenbaum and Donald~B Rubin.
\newblock The central role of the propensity score in observational studies for
  causal effects.
\newblock {\em Biometrika}, 70(1):41--55, 1983.

\bibitem{rosenfeld2020risks}
Elan Rosenfeld, Pradeep~Kumar Ravikumar, and Andrej Risteski.
\newblock The risks of invariant risk minimization.
\newblock In {\em International Conference on Learning Representations}, 2020.

\bibitem{ryu2019generalized}
Jongbin Ryu, Gitaek Kwon, Ming-Hsuan Yang, and Jongwoo Lim.
\newblock Generalized convolutional forest networks for domain generalization
  and visual recognition.
\newblock In {\em International conference on learning representations}, 2019.

\bibitem{sagawa2019distributionally}
Shiori Sagawa, Pang~Wei Koh, Tatsunori~B Hashimoto, and Percy Liang.
\newblock Distributionally robust neural networks for group shifts: On the
  importance of regularization for worst-case generalization.
\newblock {\em arXiv preprint arXiv:1911.08731}, 2019.

\bibitem{scholkopf2012causal}
Bernhard Sch{\"o}lkopf, Dominik Janzing, Jonas Peters, Eleni Sgouritsa, Kun
  Zhang, and Joris Mooij.
\newblock On causal and anticausal learning.
\newblock {\em arXiv preprint arXiv:1206.6471}, 2012.

\bibitem{scholkopf2021toward}
Bernhard Sch{\"o}lkopf, Francesco Locatello, Stefan Bauer, Nan~Rosemary Ke, Nal
  Kalchbrenner, Anirudh Goyal, and Yoshua Bengio.
\newblock Toward causal representation learning.
\newblock {\em Proceedings of the IEEE}, 109(5):612--634, 2021.

\bibitem{sand_mask}
Soroosh Shahtalebi, Jean-Christophe Gagnon-Audet, Touraj Laleh, Mojtaba
  Faramarzi, Kartik Ahuja, and Irina Rish.
\newblock Sand-mask: An enhanced gradient masking strategy for the discovery of
  invariances in domain generalization, 2021.

\bibitem{shankar2018generalizing}
Shiv Shankar, Vihari Piratla, Soumen Chakrabarti, Siddhartha Chaudhuri, Preethi
  Jyothi, and Sunita Sarawagi.
\newblock Generalizing across domains via cross-gradient training.
\newblock In {\em International Conference on Learning Representations}, 2018.

\bibitem{shi2021gradient}
Yuge Shi, Jeffrey Seely, Philip~HS Torr, N Siddharth, Awni Hannun, Nicolas
  Usunier, and Gabriel Synnaeve.
\newblock Gradient matching for domain generalization.
\newblock {\em arXiv preprint arXiv:2104.09937}, 2021.

\bibitem{sun2016deep}
Baochen Sun and Kate Saenko.
\newblock Deep coral: Correlation alignment for deep domain adaptation.
\newblock In {\em European conference on computer vision}, pages 443--450.
  Springer, 2016.

\bibitem{tobin2017domain}
Josh Tobin, Rachel Fong, Alex Ray, Jonas Schneider, Wojciech Zaremba, and
  Pieter Abbeel.
\newblock Domain randomization for transferring deep neural networks from
  simulation to the real world.
\newblock In {\em 2017 IEEE/RSJ international conference on intelligent robots
  and systems (IROS)}, pages 23--30. IEEE, 2017.

\bibitem{van2008visualizing}
Laurens Van~der Maaten and Geoffrey Hinton.
\newblock Visualizing data using t-sne.
\newblock {\em Journal of machine learning research}, 9(11), 2008.

\bibitem{vapnik1998statistical}
Vladimir Vapnik.
\newblock Statistical learning theory wiley.
\newblock {\em New York}, 1(624):2, 1998.

\bibitem{vedantam2021empirical}
Ramakrishna Vedantam, David Lopez-Paz, and David~J Schwab.
\newblock An empirical investigation of domain generalization with empirical
  risk minimizers.
\newblock {\em Advances in Neural Information Processing Systems},
  34:28131--28143, 2021.

\bibitem{venkateswara2017deep}
Hemanth Venkateswara, Jose Eusebio, Shayok Chakraborty, and Sethuraman
  Panchanathan.
\newblock Deep hashing network for unsupervised domain adaptation.
\newblock In {\em Proceedings of the IEEE conference on computer vision and
  pattern recognition}, pages 5018--5027, 2017.

\bibitem{volpi2018generalizing}
Riccardo Volpi, Hongseok Namkoong, Ozan Sener, John~C Duchi, Vittorio Murino,
  and Silvio Savarese.
\newblock Generalizing to unseen domains via adversarial data augmentation.
\newblock {\em Advances in neural information processing systems}, 31, 2018.

\bibitem{wald2021calibration}
Yoav Wald, Amir Feder, Daniel Greenfeld, and Uri Shalit.
\newblock On calibration and out-of-domain generalization.
\newblock {\em Advances in neural information processing systems},
  34:2215--2227, 2021.

\bibitem{wang2022out}
Ruoyu Wang, Mingyang Yi, Zhitang Chen, and Shengyu Zhu.
\newblock Out-of-distribution generalization with causal invariant
  transformations.
\newblock In {\em Proceedings of the IEEE/CVF Conference on Computer Vision and
  Pattern Recognition}, pages 375--385, 2022.

\bibitem{wang2022causal}
Xinyi Wang, Michael Saxon, Jiachen Li, Hongyang Zhang, Kun Zhang, and
  William~Yang Wang.
\newblock Causal balancing for domain generalization.
\newblock {\em arXiv preprint arXiv:2206.05263}, 2022.

\bibitem{wang2020heterogeneous}
Yufei Wang, Haoliang Li, and Alex~C Kot.
\newblock Heterogeneous domain generalization via domain mixup.
\newblock In {\em ICASSP 2020-2020 IEEE International Conference on Acoustics,
  Speech and Signal Processing (ICASSP)}, pages 3622--3626. IEEE, 2020.

\bibitem{wiles2021fine}
Olivia Wiles, Sven Gowal, Florian Stimberg, Sylvestre Alvise-Rebuffi, Ira
  Ktena, Taylan Cemgil, et~al.
\newblock A fine-grained analysis on distribution shift.
\newblock {\em arXiv preprint arXiv:2110.11328}, 2021.

\bibitem{xu2020adversarial}
Minghao Xu, Jian Zhang, Bingbing Ni, Teng Li, Chengjie Wang, Qi Tian, and
  Wenjun Zhang.
\newblock Adversarial domain adaptation with domain mixup.
\newblock In {\em Proceedings of the AAAI Conference on Artificial
  Intelligence}, volume~34, pages 6502--6509, 2020.

\bibitem{yan2020improve}
Shen Yan, Huan Song, Nanxiang Li, Lincan Zou, and Liu Ren.
\newblock Improve unsupervised domain adaptation with mixup training.
\newblock {\em arXiv preprint arXiv:2001.00677}, 2020.

\bibitem{yang2021adversarial}
Fu-En Yang, Yuan-Chia Cheng, Zu-Yun Shiau, and Yu-Chiang~Frank Wang.
\newblock Adversarial teacher-student representation learning for domain
  generalization.
\newblock {\em Advances in Neural Information Processing Systems},
  34:19448--19460, 2021.

\bibitem{ye2021towards}
Haotian Ye, Chuanlong Xie, Tianle Cai, Ruichen Li, Zhenguo Li, and Liwei Wang.
\newblock Towards a theoretical framework of out-of-distribution
  generalization.
\newblock {\em Advances in Neural Information Processing Systems},
  34:23519--23531, 2021.

\bibitem{ye2022ood}
Nanyang Ye, Kaican Li, Haoyue Bai, Runpeng Yu, Lanqing Hong, Fengwei Zhou,
  Zhenguo Li, and Jun Zhu.
\newblock Ood-bench: Quantifying and understanding two dimensions of
  out-of-distribution generalization.
\newblock In {\em Proceedings of the IEEE/CVF Conference on Computer Vision and
  Pattern Recognition}, pages 7947--7958, 2022.

\bibitem{yue2019domain}
Xiangyu Yue, Yang Zhang, Sicheng Zhao, Alberto Sangiovanni-Vincentelli, Kurt
  Keutzer, and Boqing Gong.
\newblock Domain randomization and pyramid consistency: Simulation-to-real
  generalization without accessing target domain data.
\newblock In {\em Proceedings of the IEEE/CVF International Conference on
  Computer Vision}, pages 2100--2110, 2019.

\bibitem{zhang2021can}
Dinghuai Zhang, Kartik Ahuja, Yilun Xu, Yisen Wang, and Aaron Courville.
\newblock Can subnetwork structure be the key to out-of-distribution
  generalization?
\newblock In {\em International Conference on Machine Learning}, pages
  12356--12367. PMLR, 2021.

\bibitem{zhang2021quantifying}
Guojun Zhang, Han Zhao, Yaoliang Yu, and Pascal Poupart.
\newblock Quantifying and improving transferability in domain generalization.
\newblock {\em Advances in Neural Information Processing Systems},
  34:10957--10970, 2021.

\bibitem{zhang2018mixup}
Hongyi Zhang, Moustapha Cisse, Yann~N Dauphin, and David Lopez-Paz.
\newblock mixup: Beyond empirical risk minimization.
\newblock In {\em International Conference on Learning Representations}, 2018.

\bibitem{zhang2018fairness}
Junzhe Zhang and Elias Bareinboim.
\newblock Fairness in decision-making—the causal explanation formula.
\newblock In {\em Proceedings of the AAAI Conference on Artificial
  Intelligence}, volume~32, 2018.

\bibitem{zhang2020adaptive}
Marvin Zhang, Henrik Marklund, Abhishek Gupta, Sergey Levine, and Chelsea Finn.
\newblock Adaptive risk minimization: A meta-learning approach for tackling
  group shift.
\newblock {\em arXiv preprint arXiv:2007.02931}, 8:9, 2020.

\bibitem{zhang2022exact}
Yabin Zhang, Minghan Li, Ruihuang Li, Kui Jia, and Lei Zhang.
\newblock Exact feature distribution matching for arbitrary style transfer and
  domain generalization.
\newblock In {\em Proceedings of the IEEE/CVF Conference on Computer Vision and
  Pattern Recognition}, pages 8035--8045, 2022.

\bibitem{zhao2020domain}
Shanshan Zhao, Mingming Gong, Tongliang Liu, Huan Fu, and Dacheng Tao.
\newblock Domain generalization via entropy regularization.
\newblock {\em Advances in Neural Information Processing Systems},
  33:16096--16107, 2020.

\bibitem{zhou2020deep}
Kaiyang Zhou, Yongxin Yang, Timothy Hospedales, and Tao Xiang.
\newblock Deep domain-adversarial image generation for domain generalisation.
\newblock In {\em Proceedings of the AAAI Conference on Artificial
  Intelligence}, volume~34, pages 13025--13032, 2020.

\bibitem{zhou2021domain}
Kaiyang Zhou, Yongxin Yang, Yu Qiao, and Tao Xiang.
\newblock Domain generalization with mixstyle.
\newblock {\em arXiv preprint arXiv:2104.02008}, 2021.

\bibitem{zhu2017unpaired}
Jun-Yan Zhu, Taesung Park, Phillip Isola, and Alexei~A Efros.
\newblock Unpaired image-to-image translation using cycle-consistent
  adversarial networks.
\newblock In {\em Proceedings of the IEEE international conference on computer
  vision}, pages 2223--2232, 2017.

\end{thebibliography}
}

\clearpage
\appendix

\section{Other Related Works}
\textbf{Learning invariant features.} To enable classifiers to generalize across domains, a very intuitive idea is to force the model to learn a representation with cross-domain invariance \cite{motiian2017unified,wald2021calibration}, which is usually implemented by adding a regularization term to the loss \cite{li2020domain,zhao2020domain,robey2021model,wang2022out,zhang2022exact}. Representative methods include using maximum mean discrepancy as a divergence measure \cite{muandet2013domain}, seeking a data representation such that the predictors using the representation are invariant (IRM) \cite{arjovsky2019invariant}, encouraging the training risks in different domains to be similar \cite{krueger2021out}, or using a correlation matrix to construct a new loss function \cite{lv2022causality}. Some works studied the conditions under which invariance can guarantee domain generalization from a theoretical point of view \cite{ye2021towards}, while other works quantified transferability of feature embeddings learned by domain generalization models \cite{zhang2021quantifying}. \cite{ahuja2020empirical} analyzed different finite sample and asymptotic behavior of ERM and IRM, and \cite{ahuja2020irmg} expanded IRM from a game theory perspective. However, existing works have demonstrated that it is difficult to achieve good generalization performance by relying only on the constraint of cross-domain invariance \cite{mahajan2021domain,ahuja2021invariance}, and some other works claimed that many of these approaches still fail to capture the invariance \cite{kamath2021does,rosenfeld2020risks}. Empirical work has also questioned the effectiveness of these methods \cite{gulrajani2020search}. On average, ERM outperforms the other methods \cite{vedantam2021empirical}.

\textbf{Data augmentation. }
Data augmentation is a kind of valid methods to impove the generalization ability of models. Domain Randomization bridge the simulated environment and the real world by generating rich enough data, which can benefit the o.o.d. generalization \cite{tobin2017domain}. Some works performed domain adversarial training to generate data across environments \cite{shankar2018generalizing,volpi2018generalizing}. Generative models and transformation models such as CycleGAN \cite{zhu2017unpaired} are also used to perform data augmentation \cite{zhou2020deep,qiao2020learning,yue2019domain}.

\textbf{Other approaches.}
In addition to the above approaches, many works have been done to enhance the performance of o.o.d. generalization in a variety of different ways. A great deal of work has been done to improve generalization performance by analyzing and designing new neural network structures \cite{li2017deeper,zhang2021can,ryu2019generalized}. Meta-learning is another helpful direction, and many approaches based on it have emerged \cite{li2018learning,balaji2018metareg,dou2019domain,li2019feature}. In addition, \cite{yang2021adversarial} proposed an approach of Adversarial Teacher-Student Representation Learning to derive generalizable representations; \cite{zhou2021domain} proposed an approach based on feature statistics mixing across source domains; \cite{piratla2020efficient} joint learned common components and domain-specific components by modifying the last classification layer; \cite{carlucci2019domain} combined supervised with self-supervised learning to improve the generalization performance of the model by solving the Jigsaw puzzles.

\section{Experimental Details}
\label{app_exp}

\subsection{Hyperparameter search}
We search hyperparameters with the same distribution as DomainBed. On DomainBed benchmark, some distribution of hyperparameters are related to image size. To avoid human intervention, we resize images of \emph{CelebA-HB} and \emph{CelebA-NS} to $224\times224$,  which is the standard size on DomainBed benchmark. To save computing resource and time, we reduce the number of search to eight for all diversity shift datasets and small-image (smaller than $224\times224$) correlation shift datasets. DRM will augment the batch, so we reduce the batchsize for big-image datasets to avoid GPU memory overflow.
The two stages of DRM use the same model: Resnet-50 \cite{he2016deep} for big-image datasets and MNIST-CNN for small-image datasets, which is consistent with DomainBed. And hyperparameters distribution of two stages are the same, which is also consistent with DomainBed. 
In stage 1, we divide the data from train environments into the training set and the validation set in a ratio of 8:2. We choose the model which perform best on the validation set.

\subsection{Datasets}
\label{ds}
\textbf{3DShapes \cite{3dshapes18}.}
To demonstrate that our approach can eliminate spurious correlations between attributes, we run our algorithm on the \emph{3DShapes} dataset. The \emph{3DShapes} is a dataset with six attributes, among which we choose the ``floor hue'' and ``orientation'' to form the spurious correlation. Specifically, we divide the orientation of the graph into two categories. Our goal is to predict which category the orientation belongs to. We use the same construction as \emph{Colored MNIST} to build three environments and add label noise. 

\textbf{DSprites \cite{dsprites17}.}
We also evaluate our DRM algorithm on the \emph{DSprites} dataset, which has six attributes. In this paper, ``Position X'' and ``Position Y'' are chosen to form spurious correlations in the \emph{DSprites}. We argue that these two attributes are similar, and thus it is challenging to identify the invariant features across all environments.

\textbf{CelebA-HB and CelebA-NS \cite{liu2015faceattributes}.}
We introduce the CelebA dataset to test the performance of our approach.
CelebA is a large-scale face attribute dataset with 40 attribute annotations, e.g., eyeglasses, wearing hat and bangs. Any two attributes can form a correlation shift dataset, one of which acts as the label to be predicted and the other is used to create spurious correlations. In this paper, for \emph{CelebA-HB}, ``No Beard'' is the label and there exists spurious correlation between the attribute ``No Beard'' and the attribute ``Wearing Hat''. For \emph{CelebA-NS}, ``Smiling" is the label and the correlation between ``Wearing Necktie'' and ``Smiling" is unstable. Unlike above mentioned datasets, correlation shift on CelebA comes from non-random sampling, which is called selection bias in causal inference. 

\subsection{Full results}
We conducted experiments on different test domains. Below are the experimental results for each domain on different datasets. We use accuracy as metrics.

For the diversity shift datasets, the support sets of data on different environments have no overlap, and the test distribution keeps the semantic content of the data unchanged while altering the data style. For instance, in the \emph{PACS} dataset, the training environment and the test environment can be photos and cartoons, respectively. We test our approach on the diversity shift datasets. Following DomainBed, we present results on RotatedMNIST \cite{ghifary2015domain}, VLCS \cite{fang2013unbiased}, PACS \cite{li2017deeper}, Office-Home \cite{venkateswara2017deep}, Terra Incognita \cite{beery2018recognition} and DomainNet \cite{peng2019moment}. In addition to the 14 methods we mentioned above, we also reports the results of other recent methods, including: Fish \cite{shi2021gradient}, Fishr \cite{rame2022fishr}, AND-mask \cite{parascandolo2020learning}, SAND-mask \cite{sand_mask}, SelfReg \cite{kim2021selfreg}, CausIRL \cite{chevalley2022invariant}, mDSDI \cite{bui2021exploiting}, SWAD \cite{cha2021swad}, and T3A \cite{iwasawa2021test}. We report the experimental results in Table \ref{table_domain_bed}. It shows that DRM does not hurt model performance for diversity shift. We observe that performance of ERM and other domain generalization algorithms are similar. The performance of our DRM approach is comparable with others on the diversity shift datasets, and improves by $\sim$$2\%$ on average concerning the DomainBed benchmark.


\begin{table}[hbtp]
\vspace{4.6cm}
\caption{The result for CMNIST}
\vspace{-1.5em}
\label{res_cm}
\begin{center}
\adjustbox{max width=\linewidth}{%
\renewcommand\arraystretch{0.75}
\begin{tabular}{lcccc}
\toprule
\textbf{Algorithm}   & \textbf{+90\%}       & \textbf{+80\%}       & \textbf{-90\%}       & \textbf{Avg}         \\
\midrule
ERM   & 71.7 $\pm$ 0.1    & 72.9 $\pm$ 0.2    & 10.0 $\pm$ 0.1    & 51.5 \\
IRM   & 72.5 $\pm$ 0.1    & 73.3 $\pm$ 0.5    & 10.2 $\pm$ 0.3    & 52.0 \\
GroupDRO    & 73.1 $\pm$ 0.3    & 73.2 $\pm$ 0.2    & 10.0 $\pm$ 0.2    & 52.1 \\
Mixup   & 72.7 $\pm$ 0.4    & 73.4 $\pm$ 0.1    & 10.1 $\pm$ 0.1    & 52.1 \\
MLDG   & 71.5 $\pm$ 0.2    & 73.1 $\pm$ 0.2    & 9.8 $\pm$ 0.1    & 51.5 \\
CORAL   & 71.6 $\pm$ 0.3    & 73.1 $\pm$ 0.1    & 9.9 $\pm$ 0.1    & 51.5 \\
MMD   & 71.4 $\pm$ 0.2    & 73.1 $\pm$ 0.2    & 9.9 $\pm$ 0.3    & 51.5 \\
DANN   & 71.4 $\pm$ 0.9    & 73.1 $\pm$ 0.1    & 10.0 $\pm$ 0.0    & 51.5 \\
CDANN   & 72.0 $\pm$ 0.2    & 73.0 $\pm$ 0.2    & 10.2 $\pm$ 0.1    & 51.7 \\
MTL   & 70.9 $\pm$ 0.2    & 72.8 $\pm$ 0.3    & 10.5 $\pm$ 0.1    & 51.4 \\
SagNet   & 71.8 $\pm$ 0.2    & 73.0 $\pm$ 0.2    & 10.3 $\pm$ 0.1    & 51.7 \\
ARM   & 82.0 $\pm$ 0.5    & 76.5 $\pm$ 0.3    & 10.2 $\pm$ 0.0    & 56.2 \\
VREx   & 72.4 $\pm$ 0.3    & 72.9 $\pm$ 0.4    & 10.2 $\pm$ 0.0    & 51.8 \\
RSC   & 71.9 $\pm$ 0.3    & 73.1 $\pm$ 0.2    & 10.0 $\pm$ 0.2    & 51.7 \\
\midrule
DRM (ours)   & 71.4 $\pm$ 0.3    & 72.4 $\pm$ 0.4    & \textbf{69.7 $\pm$ 1.5}    & \textbf{71.2} \\
\bottomrule
\end{tabular}}
\end{center}
\end{table}

\begin{table}
\vspace{-0.5cm}
\caption{The results for 3DShapes}
\vspace{-1.5em}
\begin{center}
\adjustbox{max width=\linewidth}{%
\renewcommand\arraystretch{0.75}
\begin{tabular}{lcccc}
\toprule
\textbf{Algorithm}   & \textbf{+90\%}       & \textbf{+80\%}       & \textbf{-90\%}       & \textbf{Avg}         \\
\midrule
ERM   & 74.3 ± 0.4    & 75.5 ± 0.2    & 10.1 ± 0.1    & 53.3 \\
IRM   & 74.2 ± 0.2    & 75.4 ± 0.1    & 10.0 ± 0.0    & 53.2 \\
GroupDRO    & 74.6 ± 0.1    & 75.1 ± 0.1    & 10.5 ± 0.4    & 53.4 \\
Mixup    & 74.6 ± 0.4    & 75.4 ± 0.2    & 10.2 ± 0.1    & 53.4 \\
MLDG    & 75.0 ± 0.2    & 75.4 ± 0.1    & 10.1 ± 0.1    & 53.5 \\
CORAL   & 74.6 ± 0.2    & 75.2 ± 0.2    & 10.0 ± 0.0    & 53.3 \\
MMD    & 74.6 ± 0.1    & 75.2 ± 0.1    & 10.0 ± 0.1    & 53.2 \\
DANN    & 74.6 ± 0.1    & 75.2 ± 0.1    & 10.0 ± 0.0    & 53.3 \\
CDANN    & 74.4 ± 0.4    & 75.3 ± 0.0    & 10.0 ± 0.0    & 53.3 \\
MTL    & 74.7 ± 0.2     & 75.4 ± 0.2    & 10.1 ± 0.0    & 53.4 \\
SagNet   & 74.9 ± 0.1    & 75.1 ± 0.1    & 10.1 ± 0.1    & 53.4 \\
ARM    & 81.8 ± 0.3    & 73.9 ± 0.8    & 10.0 ± 0.0    & 55.2 \\
VREx    & 74.6 ± 0.4    & 75.2 ± 0.0    & 10.8 ± 0.3    & 53.5 \\
RSC    & 74.4 ± 0.2    & 75.1 ± 0.1    & 10.1 ± 0.1    & 53.2 \\
\midrule
DRM (ours)   & 74.5 ± 0.2    & 75.1 ± 0.1    & \textbf{74.8 ± 0.1}    & \textbf{74.8} \\
\bottomrule
\end{tabular}}
\end{center}
\end{table}

\begin{table}
\caption{The results for DSprites}
\vspace{-1.5em}
\begin{center}
\adjustbox{max width=\linewidth}{%
\renewcommand\arraystretch{0.75}
\begin{tabular}{lcccc}
\toprule
\textbf{Algorithm}   & \textbf{+90\%}       & \textbf{+80\%}       & \textbf{-90\%}       & \textbf{Avg}         \\
\midrule
ERM       & 73.5 $\pm$ 0.2    & 74.8 $\pm$ 0.0    & 13.8 $\pm$ 0.5    & 54.0 \\
IRM       & 73.5 $\pm$ 0.2    & 74.2 $\pm$ 0.1    & 14.5 $\pm$ 0.3    & 54.0 \\
GroupDRO  & 73.8 $\pm$ 0.2    & 74.4 $\pm$ 0.1    & 15.0 $\pm$ 0.4    & 54.4 \\
Mixup     & 73.4 $\pm$ 0.1    & 74.3 $\pm$ 0.2    & 14.0 $\pm$ 0.3    & 53.9 \\
MLDG  & 73.9 $\pm$ 0.3    & 74.4 $\pm$ 0.3    & 14.3 $\pm$ 0.3    & 54.2 \\
CORAL     & 73.4 $\pm$ 0.3    & 74.5 $\pm$ 0.1    & 13.8 $\pm$ 0.2    & 53.9 \\
MMD       & 65.8 $\pm$ 6.4    & 74.2 $\pm$ 0.1    & 14.4 $\pm$ 0.0    & 51.4 \\
DANN      & 73.7 $\pm$ 0.5    & 73.9 $\pm$ 0.2    & 14.7 $\pm$ 0.3    & 54.1 \\
CDANN     & 73.7 $\pm$ 0.3    & 74.0 $\pm$ 0.1    & 14.4 $\pm$ 0.2    & 54.0 \\
MTL       & 73.3 $\pm$ 0.0    & 74.8 $\pm$ 0.2    & 14.8 $\pm$ 0.5    & 54.3 \\
SagNet    & 73.5 $\pm$ 0.1    & 74.8 $\pm$ 0.1    & 13.6 $\pm$ 0.1    & 54.0 \\
ARM    & 86.9 $\pm$ 0.4    & 77.6 $\pm$ 0.4    & 14.5 $\pm$ 0.6    & 59.7 \\
VREx   & 73.4 $\pm$ 0.2    & 74.6 $\pm$ 0.1    & 13.8 $\pm$ 0.3    & 53.9 \\
RSC    & 73.7 $\pm$ 0.3    & 74.5 $\pm$ 0.2    & 13.3 $\pm$ 0.2    & 53.8 \\
\midrule
DRM (ours)   & 73.7 $\pm$ 0.2    & 74.4 $\pm$ 0.1    & \textbf{73.3} $\pm$ 0.5    & \textbf{73.8} \\
\bottomrule
\end{tabular}}
\end{center}
\end{table}

\begin{table}
\caption{The results for CelebA-HB} 
\vspace{-1.5em}
\begin{center}
\adjustbox{max width=\linewidth}{%
\renewcommand\arraystretch{0.75}
\begin{tabular}{lcccc}
\toprule
\textbf{Algorithm}   & \textbf{+90\%}       & \textbf{+80\%}       & \textbf{-90\%}       & \textbf{Avg}         \\
\midrule
ERM   & 67.3  $\pm$  0.2    & 71.8 $\pm$ 0.3    & 16.8 $\pm$ 1.2    & 52.0 \\
IRM   & 65.9 $\pm$ 0.7    & 70.0 $\pm$ 0.7    & 20.4 $\pm$ 2.1    & 52.1 \\
GroupDRO    & 68.3 $\pm$ 1.2    & 71.9 $\pm$ 0.2    & 18.3 $\pm$ 1.5    & 52.8 \\
Mixup    & 68.4 $\pm$ 0.6    & 70.8 $\pm$ 0.6    & 17.9 $\pm$ 3.4    & 52.4 \\
MLDG    & 68.3 $\pm$ 0.2    & 70.6 $\pm$ 0.1    & 20.0 $\pm$ 2.1    & 53.0 \\
CORAL   & 68.3 $\pm$ 0.5    & 71.2 $\pm$ 0.2    & 17.7 $\pm$ 1.6    & 52.4 \\
MMD    & 64.5 $\pm$ 0.6    & 70.1 $\pm$ 0.6    & 17.4 $\pm$ 1.8    & 50.7 \\
DANN    & 67.0 $\pm$ 1.0    & 71.2 $\pm$ 1.5    & 16.9 $\pm$ 1.7    & 51.7 \\
CDANN    & 66.8 $\pm$ 0.8    & 72.1 $\pm$ 0.4    & 18.6 $\pm$ 2.6    & 52.5 \\
MTL    & 65.9 $\pm$ 0.9    & 71.6 $\pm$ 0.1    & 23.5 $\pm$ 1.4    & 53.7 \\
SagNet   & 64.6 $\pm$ 1.0    & 71.6 $\pm$ 0.4    & 14.9 $\pm$ 0.6    & 50.4 \\
ARM    & 66.7 $\pm$ 1.1    & 72.8 $\pm$ 0.1    & 22.8 $\pm$ 2.2    & 54.1 \\
VREx    & 66.9 $\pm$ 0.3    & 71.4 $\pm$ 0.2    & 19.2 $\pm$ 1.8    & 52.5 \\
RSC    & 67.4 $\pm$ 0.4    & 71.1 $\pm$ 0.9    & 18.9 $\pm$ 1.1    & 52.5 \\
\midrule
DRM (ours)   & 68.1 $\pm$ 1.0    & 69.3 $\pm$ 1.4    & \textbf{61.0 $\pm$ 4.9}    & \textbf{66.1} \\
\bottomrule
\end{tabular}}
\end{center}
\end{table}

\begin{table*}
\caption{The results for CelebA-NS} 
\vspace{-1.5em}
\begin{center}
\adjustbox{max width=\linewidth}{%
\renewcommand\arraystretch{0.75}
\begin{tabular}{lcccc}
\toprule
\textbf{Algorithm}   & \textbf{+90\%}       & \textbf{+80\%}       & \textbf{-90\%}       & \textbf{Avg}         \\
\midrule
ERM   & 67.8 $\pm$ 1.0    & 69.1 $\pm$ 0.6    & 21.1 $\pm$ 0.4    & 52.7 \\
IRM   & 68.2 $\pm$ 0.3    & 69.8 $\pm$ 0.1    & 21.5 $\pm$ 0.9    & 53.2 \\
GroupDRO    & 68.0 $\pm$ 0.4    & 70.6 $\pm$ 0.3    & 21.2 $\pm$ 0.2    & 53.3 \\
Mixup    & 68.7 $\pm$ 0.6    & 70.2 $\pm$ 0.7    & 22.2 $\pm$ 1.5    & 53.7 \\
MLDG    & 67.7 $\pm$ 0.3    & 70.8 $\pm$ 0.1    & 22.7 $\pm$ 1.7    & 53.7 \\
CORAL   & 68.2 $\pm$ 0.5    & 69.8 $\pm$ 0.1    & 22.1 $\pm$ 1.1    & 53.4 \\
MMD    & 68.1 $\pm$ 0.3    & 69.4 $\pm$ 0.4    & 22.5 $\pm$ 0.6    & 53.3 \\
DANN    & 69.3 $\pm$ 0.6    & 69.9 $\pm$ 0.6    & 21.8 $\pm$ 1.5    & 53.7 \\
CDANN    & 68.0 $\pm$ 0.5    & 71.1 $\pm$ 0.4    & 22.5 $\pm$ 1.2    & 53.9 \\
MTL    & 67.7 $\pm$ 0.4    & 69.5 $\pm$ 0.3    & 27.6 $\pm$ 1.2    & 54.9 \\
SagNet   & 67.8 $\pm$ 0.3    & 69.5 $\pm$ 0.2    & 22.0 $\pm$ 0.6    & 53.1 \\
ARM    & 67.7 $\pm$ 0.2    & 70.3 $\pm$ 0.3    & 21.1 $\pm$ 1.4    & 53.0 \\
VREx    &  69.7 $\pm$ 0.4    & 69.7 $\pm$ 0.3    & 20.3 $\pm$ 0.4    & 53.2 \\
RSC    & 68.9 $\pm$ 0.8    & 70.3 $\pm$ 0.2    & 23.7 $\pm$ 0.8    & 54.3 \\
\midrule
DRM (ours)   & 67.7 $\pm$ 1.1    & 68.6 $\pm$ 0.3    & \textbf{59.9 $\pm$ 2.6}    & \textbf{65.4} \\
\bottomrule
\end{tabular}}
\end{center}
\end{table*}

\begin{table*}[hbtp]
  \centering
  \tabcolsep=2.0pt
  \caption{Experimental results on the DomainBed benchmark. We use accuracy as metrics.}
  \vspace{-0.5em}
    \resizebox{1\textwidth}{!}{
    \begin{threeparttable}
    \begin{tabular}{c|c|ccccccc|c}
    \toprule
    \textbf{DomainBed} & \textbf{Algorithm} & \textbf{CMNIST} & \textbf{RMNIST} & \textbf{VLCS}  & \textbf{PACS}  & \textbf{OfficeHome} & \textbf{TerraInc} & \textbf{DomainNet} & \textbf{Avg} \\
    \midrule
    \multirow{14}{*}{Official report} &
    ERM   & 51.5 ± 0.1 & 98.0 ± 0.0 & 77.5 ± 0.4 & 85.5 ± 0.2 & 66.5 ± 0.3 & 46.1 ± 1.8 & 40.9 ± 0.1 & 66.6 \\
    & IRM   & 52.0 ± 0.1 & 97.7 ± 0.1 & 78.5 ± 0.5 & 83.5 ± 0.8 & 64.3 ± 2.2 & 47.6 ± 0.8 & 33.9 ± 2.8 & 65.4 \\
    & GroupDRO & 52.1 ± 0.0 & 98.0 ± 0.0 & 76.7 ± 0.6 & 84.4 ± 0.8 & 66.0 ± 0.7 & 43.2 ± 1.1 & 33.3 ± 0.2 & 64.8 \\
    & Mixup & 52.1 ± 0.2 & 98.0 ± 0.1 & 77.4 ± 0.6  & 84.6 ± 0.6 & 68.1 ± 0.3 & 47.9 ± 0.8 & 39.2 ± 0.1 & 66.7 \\
    & MLDG  & 51.5 ± 0.1 & 97.9 ± 0.0 & 77.2 ± 0.4 & 84.9 ± 1.0 & 66.8 ± 0.6 & 47.7 ± 0.9 & 41.2 ± 0.1 & 66.7 \\
    & CORAL & 51.5 ± 0.1 & 98.0 ± 0.1 & 78.8 ± 0.6 & 86.2 ± 0.3 & 68.7 ± 0.3 & 47.6 ± 1.0 & 41.5 ± 0.1 & 67.5 \\
    & MMD   & 51.5 ± 0.2 & 97.9 ± 0.0 & 77.5 ± 0.9 & 84.6 ± 0.5 & 66.3 ± 0.1 & 42.2 ± 1.6 & 23.4 ± 9.5 & 63.3 \\
    & DANN  & 51.5 ± 0.3 & 97.8 ± 0.1 & 78.6 ± 0.4 & 83.6 ± 0.4 & 65.9 ± 0.6 & 46.7 ± 0.5 & 38.3 ± 0.1 & 66.1 \\
    & CDANN & 51.7 ± 0.1 & 97.9 ± 0.1 & 77.5 ± 0.1 & 82.6 ± 0.9 & 65.8 ± 1.3 & 45.8 ± 1.6 & 38.3 ± 0.3 & 65.6 \\
    & MTL   & 51.4 ± 0.1 & 97.9 ± 0.0 & 77.2 ± 0.4 & 84.6 ± 0.5 & 66.4 ± 0.5 & 45.6 ± 1.2 & 40.6 ± 0.1 & 66.2 \\
    & SagNet & 51.7 ± 0.0 & 98.0 ± 0.0 & 77.8 ± 0.5 & 86.3 ± 0.2 & 68.1 ± 0.1 & 48.6 ± 1.0 & 40.3 ± 0.1 & 67.2 \\
    & ARM   & 56.2 ± 0.2 & 98.2 ± 0.1 & 77.6 ± 0.3 & 85.1 ± 0.4 & 64.8 ± 0.3 & 45.5 ± 0.3 & 35.5 ± 0.2 & 66.1 \\
    & VREx  & 51.8 ± 0.1 & 97.9 ± 0.1 & 78.3 ± 0.2 & 84.9 ± 0.6 & 66.4 ± 0.6 & 46.4 ± 0.6 & 33.6 ± 2.9 & 65.6 \\
    & RSC   & 51.7 ± 0.2 & 97.6 ± 0.1 & 77.1 ± 0.5 & 85.2 ± 0.9 & 65.5 ± 0.9 & 46.6 ± 1.0 & 38.9 ± 0.5 & 66.1 \\
    \midrule
    \multirow{7}{*}{Codes by authors} & Fish  & 51.6 ± 0.1 & 98.0 ± 0.0 & 77.8 ± 0.3 & 85.5 ± 0.3 & 68.6 ± 0.4 & 45.1 ± 1.3 & 42.7 ± 0.2 & 67.1 \\
    & Fishr & 52.0 ± 0.2 & 97.8 ± 0.0 & 77.8 ± 0.1 & 85.5 ± 0.4 & 67.8 ± 0.1 & 47.4 ± 1.6 & 41.7 ± 0.0 & 67.1 \\
    & ANDmask & 51.3 ± 0.2 & 97.6 ± 0.1 & 78.1 ± 0.9 & 84.4 ± 0.9 & 65.6 ± 0.4 & 44.6 ± 0.3 & 37.2 ± 0.6 & 65.5 \\
    & SANDmask & 51.8 ± 0.2 & 97.4 ± 0.1 & 77.4 ± 0.2 & 84.6 ± 0.9 & 65.8 ± 0.4 & 42.9 ± 1.7 & 32.1 ± 0.6 & 64.6 \\
    & SelfReg & 52.1 ± 0.2 & 98.0 ± 0.1 & 77.8 ± 0.9 & 85.6 ± 0.4 & 67.9 ± 0.7 & 47.0 ± 0.3 & 42.8 ± 0.0 & 67.3 \\
    & CausIRL\_CORAL & 51.7 ± 0.1 & 97.9 ± 0.1 & 77.5 ± 0.6 & 85.8 ± 0.1 & 68.6 ± 0.3 & 47.3 ± 0.8 & 41.9 ± 0.1 & 67.3 \\
    & CausIRL\_MMD & 51.6 ± 0.1 & 97.9 ± 0.0 & 77.6 ± 0.4 & 84.0 ± 0.8 & 65.7 ± 0.6 & 46.3 ± 0.9 & 40.3 ± 0.2 & 66.2 \\
    \midrule
    \multirow{3}{*}{Reported by authors} 
    & mDSDI & 52.2 ± 0.2 & 98.0 ± 0.1 & 79.0 ± 0.3 & 86.2 ± 0.2 & 69.2 ± 0.4 & 48.1 ± 1.4 & 42.8 ± 0.1 & 67.9 \\
    & SWAD & - & - & 79.1 ± 0.1 & 88.1 ± 0.1 & 70.6 ± 0.2 & 50.0 ± 0.3 & 46.5 ± 0.1 & - \\
    & T3A & - & - & 80.0 ± 0.2 & 85.3 ± 0.6 & 68.3 ± 0.1 & 47.0 ± 0.6 & - & - \\
    \midrule
    & DRM (ours) & \textbf{71.2} ± 0.6 & 97.6 ± 0.1 & 77.9 ± 0.5 & 84.8 ± 0.5 & 65.7 ± 0.6 & 48.2 ± 0.2 &  41.0 ± 0.2 &  \textbf{69.5} \\
    \bottomrule
    \end{tabular}%
    
    \end{threeparttable}
    }
  \label{table_domain_bed}%
\end{table*}%

\begin{table*}[htbp]
\caption{The result for RMNIST} 
\vspace{-1.5em}
\begin{center}
\adjustbox{max width=\textwidth}{%
\renewcommand\arraystretch{0.75}
\begin{tabular}{lccccccc}
\toprule
\textbf{Algorithm}   & \textbf{0}  & \textbf{15}  & \textbf{30}  & \textbf{45}   & \textbf{60}    & \textbf{75}   & \textbf{Avg}   \\
\midrule
    ERM   & 95.9 $\pm$ 0.1   & 98.9 $\pm$ 0.0  & 98.8 $\pm$ 0.0  & 98.9 $\pm$ 0.0  & 98.9 $\pm$ 0.0   & 96.4 $\pm$ 0.0  & 98.0 \\
    IRM   & 95.5 $\pm$ 0.1   & 98.8 $\pm$ 0.2  & 98.7 $\pm$ 0.1  & 98.6 $\pm$ 0.1  & 98.7 $\pm$ 0.0   & 95.9 $\pm$ 0.2  & 97.7 \\
    GroupDRO & 95.6 $\pm$ 0.1   & 98.9 $\pm$ 0.1  & 98.9 $\pm$ 0.1  & 99.0 $\pm$ 0.0  & 98.9 $\pm$ 0.0   & 96.5 $\pm$ 0.2  & 98.0 \\
    Mixup & 95.8 $\pm$ 0.3   & 98.9 $\pm$ 0.0  & 98.9 $\pm$ 0.0  & 98.9 $\pm$ 0.0  & 98.8 $\pm$ 0.1   & 96.5 $\pm$ 0.3  & 98.0 \\
    MLDG  & 95.8 $\pm$ 0.1   & 98.9 $\pm$ 0.1  & 99.0 $\pm$ 0.0  & 98.9 $\pm$ 0.1  & 99.0 $\pm$ 0.0   & 95.8 $\pm$ 0.3  & 97.9 \\
    CORAL & 95.8 $\pm$ 0.3   & 98.8 $\pm$ 0.0  & 98.9 $\pm$ 0.0  & 99.0 $\pm$ 0.0  & 98.9 $\pm$ 0.1   & 96.4 $\pm$ 0.2  & 98.0 \\
    MMD   & 95.6 $\pm$ 0.1   & 98.9 $\pm$ 0.1  & 99.0 $\pm$ 0.0  & 99.0 $\pm$ 0.0  & 98.9 $\pm$ 0.0   & 96.0 $\pm$ 0.2  & 97.9 \\
    DANN  & 95.0 $\pm$ 0.5   & 98.9 $\pm$ 0.1  & 99.0 $\pm$ 0.0  & 90.0 $\pm$ 0.1  & 98.9 $\pm$ 0.0   & 96.3 $\pm$ 0.2  & 97.8 \\
    CDANN & 95.7 $\pm$ 0.2   & 98.8 $\pm$ 0.0  & 98.9 $\pm$ 0.1  & 98.9 $\pm$ 0.1  & 98.9 $\pm$ 0.1   & 96.1 $\pm$ 0.3  & 97.9 \\
    MTL   & 95.6 $\pm$ 0.1   & 99.0 $\pm$ 0.1  & 99.0 $\pm$ 0.0  & 98.9 $\pm$ 0.1  & 99.0 $\pm$ 0.1   & 95.8 $\pm$ 0.2  & 97.9 \\
    SagNet & 95.9 $\pm$ 0.3   & 98.9 $\pm$ 0.1  & 99.0 $\pm$ 0.1  & 99.1 $\pm$ 0.0  & 99.0 $\pm$ 0.1   & 96.3 $\pm$ 0.1  & 98.0 \\
    ARM   & 96.7 $\pm$ 0.2   & 99.1 $\pm$ 0.0  & 99.0 $\pm$ 0.0  & 99.0 $\pm$ 0.1  & 99.1 $\pm$ 0.1   & 96.5 $\pm$ 0.4  & 98.2 \\
    VREx  & 95.9 $\pm$ 0.2   & 99.0 $\pm$ 0.1  & 98.9 $\pm$ 0.1  & 98.9 $\pm$ 0.1  & 98.7 $\pm$ 0.1   & 96.2 $\pm$ 0.2  & 97.9 \\
    RSC   & 94.8 $\pm$ 0.5   & 98.7 $\pm$ 0.1  & 98.8 $\pm$ 0.1  & 98.8 $\pm$ 0.0  & 98.9 $\pm$ 0.1   & 95.9 $\pm$ 0.2  & 97.6 \\
\midrule
    DRM (ours)  & 94.5 $\pm$ 0.6   & 98.6 $\pm$ 0.1  & 98.8 $\pm$ 0.1  & 99.1 $\pm$ 0.0  & 98.9 $\pm$ 0.0   & 96.0 $\pm$ 0.3 & 97.6  \\
\bottomrule
\end{tabular}}
\end{center}
\end{table*}

\begin{table*}
\caption{The result for VLCS} 
\vspace{-1.5em}
\begin{center}
\adjustbox{max width=\textwidth}{%
\renewcommand\arraystretch{0.75}
\begin{tabular}{lccccc}
\toprule
\textbf{Algorithm}   & \textbf{C}  & \textbf{L}  & \textbf{S}  & \textbf{V}   & \textbf{Avg}   \\
\midrule
    ERM   & 97.7 $\pm$ 0.4   & 64.3 $\pm$ 0.9  & 73.4 $\pm$ 0.5  & 74.6 $\pm$ 1.3  & 77.5 \\
    IRM   & 98.6 $\pm$ 0.1   & 64.9 $\pm$ 0.9  & 73.4 $\pm$ 0.6  & 77.3 $\pm$ 0.9  & 78.5 \\
    GroupDRO & 97.3 $\pm$ 0.3  & 63.4 $\pm$ 0.9  & 69.5 $\pm$ 0.8  & 76.7 $\pm$ 0.7  & 76.7 \\
    Mixup & 98.3 $\pm$ 0.6  & 64.8 $\pm$ 1.0  & 72.1 $\pm$ 0.5  & 74.3 $\pm$ 0.8  & 77.4 \\
    MLDG  & 97.4 $\pm$ 0.2  & 65.2 $\pm$ 0.7  & 71.0 $\pm$ 1.4  & 75.3 $\pm$ 1.0  & 77.2 \\
    CORAL & 98.3 $\pm$ 0.1  & 66.1 $\pm$ 1.2    & 73.4 $\pm$ 0.3  & 77.5 $\pm$ 1.2  & 78.8 \\
    MMD   & 97.7 $\pm$ 0.1  & 64.0 $\pm$ 1.1  & 72.8 $\pm$ 0.2  & 75.3 $\pm$ 3.3  & 77.5 \\
    DANN  & 99.0 $\pm$ 0.3  & 65.1 $\pm$ 1.4  & 73.1 $\pm$ 0.3  & 77.2 $\pm$ 0.6  & 78.6 \\
    CDANN & 97.1 $\pm$ 0.3  & 65.1 $\pm$ 1.2  & 70.7 $\pm$ 0.8  & 77.1 $\pm$ 1.5  & 77.5 \\
    MTL   & 97.8 $\pm$ 0.4  & 64.3 $\pm$ 0.3  & 71.5 $\pm$ 0.7  & 75.3 $\pm$ 1.7  & 77.2 \\
    SagNet & 97.9 $\pm$ 0.4  & 64.5 $\pm$ 0.5  & 71.4 $\pm$ 1.3  & 77.5 $\pm$ 0.5    & 77.8 \\
    ARM   & 98.7 $\pm$ 0.2  & 63.6 $\pm$ 0.7  & 71.3 $\pm$ 1.2  & 76.7 $\pm$ 0.6  & 77.6 \\
    VREx  & 98.4 $\pm$ 0.3    & 64.4 $\pm$ 1.4  & 74.1 $\pm$ 0.4  & 76.2 $\pm$ 1.3  & 78.3 \\
    RSC   & 97.9 $\pm$ 0.2  & 64.4 $\pm$ 1.4  & 74.1 $\pm$ 0.4  & 76.2 $\pm$ 1.3  & 77.1 \\
\midrule
    DRM (ours)  & 97.9 $\pm$ 0.2  & 65.1 $\pm$ 0.7  & 71.5 $\pm$ 0.9  & 77.1 $\pm$ 1.7  &  77.9\\

\bottomrule
\end{tabular}}
\end{center}
\end{table*}

\begin{table*}[htbp]
\caption{The result for PACS} 
\vspace{-1.5em}
\begin{center}
\adjustbox{max width=\textwidth}{%
\renewcommand\arraystretch{0.75}
\begin{tabular}{lccccc}
\toprule
\textbf{Algorithm}   & \textbf{A}  & \textbf{C}  & \textbf{P}  & \textbf{S}   & \textbf{Avg}   \\
\midrule
    ERM   & 84.7 $\pm$ 0.4   & 80.8 $\pm$ 0.6  & 97.2 $\pm$ 0.3  & 79.3 $\pm$ 1.0  & 85.5 \\
    IRM   & 84.8 $\pm$ 1.3   & 76.4 $\pm$ 1.1  & 96.7 $\pm$ 0.6  & 76.1 $\pm$ 1.0  & 83.5 \\
    GroupDRO & 83.5 $\pm$ 0.9  & 79.1 $\pm$ 0.6  & 96.7 $\pm$ 0.3  & 78.3 $\pm$ 2.0  & 84.4 \\
    Mixup & 86.1 $\pm$ 0.5  & 78.9 $\pm$ 0.8  & 97.6 $\pm$ 0.1  & 75.8 $\pm$ 1.8  & 84.6 \\
    MLDG  & 85.5 $\pm$ 1.4  & 80.1 $\pm$ 1.7  & 97.4 $\pm$ 0.3  & 76.6 $\pm$ 1.1  & 84.9 \\
    CORAL & 88.3 $\pm$ 0.2  & 80 $\pm$ 0.5    & 97.5 $\pm$ 0.3  & 78.8 $\pm$ 1.3  & 86.2 \\
    MMD   & 86.1 $\pm$ 1.4  & 79.4 $\pm$ 0.9  & 96.6 $\pm$ 0.2  & 76.5 $\pm$ 0.5  & 84.6 \\
    DANN  & 86.4 $\pm$ 0.8  & 77.4 $\pm$ 0.8  & 97.3 $\pm$ 0.4  & 73.5 $\pm$ 2.3  & 83.6 \\
    CDANN & 84.6 $\pm$ 1.8  & 75.5 $\pm$ 0.9  & 96.8 $\pm$ 0.3  & 73.5 $\pm$ 0.6  & 82.6 \\
    MTL   & 87.5 $\pm$ 0.8  & 77.1 $\pm$ 0.5  & 96.4 $\pm$ 0.8  & 77.3 $\pm$ 1.8  & 84.6 \\
    SagNet & 87.4 $\pm$ 1.0  & 80.7 $\pm$ 0.6  & 97.1 $\pm$ 0.1  & 80 $\pm$ 0.4    & 86.3 \\
    ARM   & 86.8 $\pm$ 0.6  & 76.8 $\pm$ 0.5  & 97.4 $\pm$ 0.3  & 79.3 $\pm$ 1.2  & 85.1 \\
    VREx  & 86 $\pm$ 1.6    & 79.1 $\pm$ 0.6  & 96.9 $\pm$ 0.5  & 77.7 $\pm$ 1.7  & 84.9 \\
    RSC   & 85.4 $\pm$ 0.8  & 79.7 $\pm$ 1.8  & 97.6 $\pm$ 0.3  & 78.2 $\pm$ 1.2  & 85.2 \\
\midrule
    DRM (ours)  & 85.0 $\pm$ 0.9  & 80.0 $\pm$ 0.5  & 96.7 $\pm$ 0.6  & 77.5 $\pm$ 1.2  &  84.8\\

\bottomrule
\end{tabular}}
\end{center}
\end{table*}

\begin{table*}
\caption{The result for OFFICEHOME} 
\vspace{-1.5em}
\begin{center}
\adjustbox{max width=\textwidth}{%
\renewcommand\arraystretch{0.75}
\begin{tabular}{lccccc}
\toprule
\textbf{Algorithm}   & \textbf{A}  & \textbf{C}  & \textbf{P}  & \textbf{R}   & \textbf{Avg}   \\
\midrule
    ERM   & 61.3 $\pm$ 0.7   & 52.4 $\pm$ 0.3  & 75.8 $\pm$ 0.1  & 76.6 $\pm$ 0.3  & 66.5 \\
    IRM   & 58.9 $\pm$ 2.3   & 52.2 $\pm$ 1.6  & 72.1 $\pm$ 2.9  & 74.0 $\pm$ 2.5  & 64.3 \\
    GroupDRO & 60.4 $\pm$ 0.7  & 52.7 $\pm$ 1.0  & 75.0 $\pm$ 0.7  & 76.0 $\pm$ 0.7  & 66.0 \\
    Mixup & 62.4 $\pm$ 0.8  & 54.8 $\pm$ 0.6  & 76.9 $\pm$ 0.3  & 78.3 $\pm$ 0.2  & 68.1 \\
    MLDG  & 61.5 $\pm$ 0.9  & 53.2 $\pm$ 0.6  & 75.0 $\pm$ 1.2  & 77.5 $\pm$ 0.4  & 66.8 \\
    CORAL & 65.3 $\pm$ 0.4  & 54.4 $\pm$ 0.5    & 76.5 $\pm$ 0.1  & 78.4 $\pm$ 0.5  & 68.7 \\
    MMD   & 60.4 $\pm$ 0.2  & 53.3 $\pm$ 0.3  & 74.3 $\pm$ 0.1  & 77.4 $\pm$ 0.6  & 66.3 \\
    DANN  & 59.9 $\pm$ 1.3  & 53.0 $\pm$ 0.3  & 73.6 $\pm$ 0.7  & 76.9 $\pm$ 0.5  & 65.9 \\
    CDANN & 61.5 $\pm$ 1.4  & 50.4 $\pm$ 2.4  & 74.4 $\pm$ 0.9  & 76.6 $\pm$ 0.8  & 65.8 \\
    MTL   & 61.5 $\pm$ 0.7  & 52.4 $\pm$ 0.6  & 74.9 $\pm$ 0.4  & 76.8 $\pm$ 0.4  & 66.4 \\
    SagNet & 63.4 $\pm$ 0.2  & 54.8 $\pm$ 0.4  & 75.8 $\pm$ 0.4  & 78.3 $\pm$ 0.3    & 68.1 \\
    ARM   & 58.9 $\pm$ 0.8  & 51.0 $\pm$ 0.5  & 74.1 $\pm$ 0.1  & 75.2 $\pm$ 0.3  & 64.8 \\
    VREx  & 60.7 $\pm$ 0.9    & 53.0 $\pm$ 0.9  & 75.3 $\pm$ 0.1  & 76.6 $\pm$ 0.5  & 66.4 \\
    RSC   & 60.7 $\pm$ 1.4  & 51.4 $\pm$ 0.3  & 74.8 $\pm$ 1.1  & 75.1 $\pm$ 1.3  & 65.5 \\
\midrule
    DRM (ours)  & 60.4 $\pm$ 0.6  & 52.5 $\pm$ 0.5  & 74.2 $\pm$ 0.6  & 75.5 $\pm$ 0.9  &  65.7\\

\bottomrule
\end{tabular}}
\end{center}
\end{table*}

\begin{table*}
\caption{The result for TERRAINCOGNITA} 
\vspace{-1.5em}
\begin{center}
\adjustbox{max width=\textwidth}{%
\renewcommand\arraystretch{0.75}
\begin{tabular}{lccccc}
\toprule
\textbf{Algorithm}   & \textbf{L100}  & \textbf{L38}  & \textbf{L43}  & \textbf{L46}   & \textbf{Avg}   \\
\midrule
    ERM   & 49.8 $\pm$ 4.4   & 42.1 $\pm$ 1.4  & 56.9 $\pm$ 1.8  & 35.7 $\pm$ 3.9  & 46.1 \\
    IRM   & 54.6 $\pm$ 1.3   & 39.8 $\pm$ 1.9  & 56.2 $\pm$ 1.8  & 39.6 $\pm$ 0.8  & 47.6 \\
    GroupDRO & 41.2 $\pm$ 0.7  & 38.6 $\pm$ 2.1  & 56.7 $\pm$ 0.9  & 36.4 $\pm$ 2.1  & 43.2 \\
    Mixup & 59.6 $\pm$ 2.0  & 42.2 $\pm$ 1.4  & 55.9 $\pm$ 0.8  & 33.9 $\pm$ 1.4  & 47.9 \\
    MLDG  & 54.2 $\pm$ 3.0  & 44.3 $\pm$ 1.1  & 55.6 $\pm$ 0.3  & 36.9 $\pm$ 2.2  & 47.7 \\
    CORAL & 51.6 $\pm$ 2.4  & 42.2 $\pm$ 1.0    & 57.0 $\pm$ 1.0  & 39.8 $\pm$ 2.9  & 47.6 \\
    MMD   & 41.9 $\pm$ 3.0  & 34.8 $\pm$ 1.0  & 57.0 $\pm$ 1.9  & 35.2 $\pm$ 1.8  & 42.2 \\
    DANN  & 51.1 $\pm$ 3.5  & 40.6 $\pm$ 0.6  & 57.4 $\pm$ 0.5  & 37.7 $\pm$ 1.8  & 46.7 \\
    CDANN & 47.0 $\pm$ 1.9  & 41.3 $\pm$ 4.8  & 54.9 $\pm$ 1.7  & 39.8 $\pm$ 2.3  & 45.8 \\
    MTL   & 49.3 $\pm$ 1.2  & 39.6 $\pm$ 6.3  & 55.6 $\pm$ 1.1  & 37.8 $\pm$ 0.8  & 45.6 \\
    SagNet & 53.0 $\pm$ 2.9  & 43.0 $\pm$ 2.5  & 57.9 $\pm$ 0.6  & 40.4 $\pm$ 1.3    & 48.6 \\
    ARM   & 49.3 $\pm$ 0.7  & 38.3 $\pm$ 2.4  & 55.8 $\pm$ 0.8  & 38.7 $\pm$ 1.3  & 45.5 \\
    VREx  & 48.2 $\pm$ 4.3    & 41.7 $\pm$ 1.3  & 56.8 $\pm$ 0.8  & 38.7 $\pm$ 3.1  & 46.4 \\
    RSC   & 50.2 $\pm$ 2.2  & 39.2 $\pm$ 1.4  & 56.3 $\pm$ 1.4  & 40.8 $\pm$ 0.6  & 46.6 \\
\midrule
    DRM (ours)  & 52.8 $\pm$ 3.6  & 42.7 $\pm$ 1.3  & 56.3 $\pm$ 1.2  & 41.1 $\pm$ 2.0  &  48.2\\

\bottomrule
\end{tabular}}
\end{center}
\end{table*}

\begin{table*}[htbp]
  \centering
  \caption{The result for DOMAINNET}
  \vspace{-0.5em}
  \renewcommand\arraystretch{0.75}
    \begin{tabular}{lccccccc}
    \toprule
    \textbf{Algorithm} & \textbf{clip}  & \textbf{info}  & \textbf{paint} & \textbf{quick} & \textbf{real} & \textbf{sketch} & \textbf{Avg} \\
    \midrule
    ERM   & 58.1 $\pm$ 0.3  & 18.8 $\pm$ 0.3  & 16.7 $\pm$ 0.3  & 12.2 $\pm$ 0.4  & 59.6 $\pm$ 0.1  & 49.8 $\pm$ 0.4 & 40.9 \\
    IRM   & 48.5 $\pm$ 2.8  & 15.0 $\pm$ 1.5  & 38.3 $\pm$ 4.3  & 10.9 $\pm$ 0.5  & 48.2 $\pm$ 5.2  & 42.3 $\pm$ 3.1 & 33.9 \\
    GroupDRO & 47.2 $\pm$ 0.5  & 17.5 $\pm$ 0.4  & 33.8 $\pm$ 0.5  & 9.3 $\pm$ 0.3  & 51.6 $\pm$ 0.4  & 40.1 $\pm$ 0.6 & 33.3 \\
    Mixup & 55.7 $\pm$ 0.3  & 18.5 $\pm$ 0.5  & 44.3 $\pm$ 0.5  & 12.5 $\pm$ 0.4  & 55.8 $\pm$ 0.1  & 48.2 $\pm$ 0.5 & 39.2 \\
    MLDG  & 59.1 $\pm$ 0.2  & 19.1 $\pm$ 0.3  & 45.8 $\pm$ 0.7  & 13.4 $\pm$ 0.3  & 59.6 $\pm$ 0.2  & 50.2 $\pm$ 0.4 & 41.2 \\
    CORAL & 59.2 $\pm$ 0.1  & 19.7 $\pm$ 0.2  & 46.6 $\pm$ 0.3  & 13.4 $\pm$ 0.3  & 59.8 $\pm$ 0.2  & 50.1 $\pm$ 0.6 & 41.5 \\
    MMD   & 32.1 $\pm$ 13.3  & 11.0 $\pm$ 4.6  & 26.8 $\pm$ 11.3  & 8.7 $\pm$ 2.1  & 32.7 $\pm$ 13.8  & 28.9 $\pm$ 11.9 & 23.4 \\
    DANN  & 53.1 $\pm$ 0.2  & 18.3 $\pm$ 0.1  & 44.2 $\pm$ 0.7  & 11.8 $\pm$ 0.1  & 55.5 $\pm$ 0.4  & 46.8 $\pm$ 0.6 & 38.3 \\
    CDANN & 54.6 $\pm$ 0.4  & 17.3 $\pm$ 0.1  & 43.7 $\pm$ 0.9  & 12.1 $\pm$ 0.7  & 56.2 $\pm$ 0.4  & 45.9 $\pm$ 0.5 & 38.3 \\
    MTL   & 57.9 $\pm$ 0.5  & 18.5 $\pm$ 0.4  & 46.0 $\pm$ 0.1  & 12.5 $\pm$ 0.1  & 59.5 $\pm$ 0.3  & 49.2 $\pm$ 0.1 & 40.6 \\
    SagNet & 57.7 $\pm$ 0.3  & 19.0 $\pm$ 0.2  & 45.3 $\pm$ 0.3  & 12.7 $\pm$ 0.5  & 58.1 $\pm$ 0.5  & 48.8 $\pm$ 0.2 & 40.3 \\
    ARM   & 49.7 $\pm$ 0.3  & 16.3 $\pm$ 0.5  & 40.9 $\pm$ 1.1  & 9.4 $\pm$ 0.1  & 53.4 $\pm$ 0.4  & 43.5 $\pm$ 0.4 & 35.5 \\
    VREx  & 47.3 $\pm$ 3.5  & 16.0 $\pm$ 1.5  & 35.8 $\pm$ 4.6  & 10.9 $\pm$ 0.3  & 49.6 $\pm$ 4.9  & 42.0 $\pm$ 3.0 & 33.6\\
    RSC   & 55.0 $\pm$ 1.2  & 18.3 $\pm$ 0.5  & 44.4 $\pm$ 0.6  & 12.2 $\pm$ 0.2  & 55.7 $\pm$ 0.7  & 47.8 $\pm$ 0.9 & 38.9 \\
    \midrule
    DRM (ours) & 58.5 $\pm$ 0.5  & 19.5 $\pm$ 0.4  & 45.4 $\pm$ 0.1  & 13.8 $\pm$ 0.6  & 59.0 $\pm$ 1.0  & 49.9 $\pm$ 0.7 & 41.0\\
    \bottomrule
    \end{tabular}%
  
\end{table*}%

\end{document}